\documentclass[11pt]{article}

\usepackage[preprint]{acl}

\usepackage{times}
\usepackage{latexsym}

\usepackage[T1]{fontenc}

\usepackage[utf8]{inputenc}

\usepackage{microtype}

\usepackage{inconsolata}

\usepackage{graphicx}

\usepackage{booktabs}
\usepackage{multirow}
\usepackage{enumitem}
\usepackage{amssymb}
\usepackage{amsmath}
\usepackage{algorithm}
\usepackage{algpseudocode}
\usepackage{booktabs}
\usepackage{array}

\usepackage[most]{tcolorbox}

\usepackage[most]{tcolorbox}
\usepackage{listings}
\usepackage{xcolor}
\usepackage{bbm}

\usepackage{acl}

\tcbuselibrary{breakable,skins,listings}

\newtcolorbox{promptbox}{
    enhanced jigsaw,
    breakable,
    colback=gray!4,
    colframe=black!50,
    boxrule=0.45pt,
    arc=0.8mm,
    outer arc=0.8mm,
    boxsep=0pt,
    left=1.4mm,
    right=1.4mm,
    top=1.0mm,
    bottom=1.0mm,
    before skip=5pt plus 1pt minus 1pt,
    after skip=5pt plus 1pt minus 1pt,
    fontupper=\small,
    before upper={%
        \raggedright
        \setlength{\parindent}{0pt}%
        \setlength{\parskip}{0.35em}%
        \setlength{\emergencystretch}{2em}%
    },
}

\lstdefinestyle{promptjson}{
    basicstyle=\ttfamily\footnotesize,
    columns=fullflexible,
    keepspaces=true,
    showstringspaces=false,
    breaklines=true,
    breakatwhitespace=false,
    upquote=true,
    tabsize=2,
    aboveskip=0pt,
    belowskip=0pt,
    xleftmargin=0pt,
    xrightmargin=0pt,
}

\newtcblisting{promptcode}{
    enhanced,
    listing only,
    listing engine=listings,
    listing options={style=promptjson},
    colback=white,
    colframe=black!18,
    boxrule=0.3pt,
    arc=0.5mm,
    boxsep=0pt,
    left=0.8mm,
    right=0.8mm,
    top=0.6mm,
    bottom=0.6mm,
    before skip=2pt,
    after skip=1pt,
}

\newcommand{\promptsec}[1]{%
    \par\smallskip
    \noindent\textbf{#1}\par\nobreak
}

\newcommand{\placeholder}[1]{%
    \texttt{\textless\detokenize{#1}\textgreater}%
}

\title{AirAnchor: Bridging Local and Global Spatial Information for Zero-Shot Aerial Vision-and-Language Navigation}

\author{
  \textbf{Shanwei Fan\textsuperscript{1,2}},
  \textbf{Bin Zhang\textsuperscript{1,2}},
  \textbf{Zhiwei Xu\textsuperscript{3}},
  \textbf{Yingxuan Teng\textsuperscript{1,2}}
  \\
  \textbf{Siqi Dai\textsuperscript{1,2}},
  \textbf{Lin Cheng\textsuperscript{1,2}},
  \textbf{Guoliang Fan\textsuperscript{1,2}}
  \\
  \\
  \textsuperscript{1}National Key Laboratory of Cognition and Decision Intelligence for Complex Systems,\\
  Institute of Automation, Chinese Academy of Sciences, Beijing, China
  \\
  \textsuperscript{2}School of Artificial Intelligence, University of Chinese Academy of Sciences, Beijing, China
  \\
  \textsuperscript{3}School of Artificial Intelligence, Shandong University, Jinan, Shandong, China
  \\
  \\
  {\small
  \texttt{\{fanshanwei2024,zhangbin2020,tengyingxuan2024\}@ia.ac.cn}}
  \\
  {\small
  \texttt{\{daisiqi2025,chenglin2025,guoliang.fan\}@ia.ac.cn}
  \quad
  \texttt{zhiwei\_xu@sdu.edu.cn}}
}

\begin{document}
\maketitle
\begin{abstract}
Aerial Vision-and-Language Navigation requires drones to follow natural-language instructions and navigate through complex urban environments. Accurate navigation relies on both local and global spatial information, which support immediate action grounding and long-horizon path planning, respectively. However, existing zero-shot methods typically operate at a single spatial scale, relying either on local representations constructed online from current observations or on global memories built offline from historical experience. To address this limitation, we propose AirAnchor, a new paradigm that bridges local and global spatial information through spatial anchors and integrates both into a shared navigation framework, enabling comprehensive spatial grounding for decision-making. AirAnchor consists of three core components: (1) Query-Driven Spatial Anchor Grounding, which identifies decision-relevant anchors from visual observations and organizes them into local spatial representations; (2) Persistent Object Spatial Memory, which incrementally maintains an object knowledge base as persistent global spatial memory and retrieves landmark-related spatial priors; and (3) a Spatially-Informed Navigation Agent, which explicitly integrates both local and global spatial information into an agentic framework for decision-making. Extensive experiments on \textit{AerialVLN} demonstrate that AirAnchor substantially outperforms existing zero-shot baselines, validating the effectiveness and efficiency of the proposed paradigm.
\end{abstract}

\section{Introduction}
Aerial Vision-and-Language Navigation (Aerial VLN)~\cite{liu2023aerialvln, gao2025openfly, wang2025towards, NEURIPS2025_92cfa104} has emerged as a novel and challenging embodied AI task that requires an unmanned aerial vehicle (UAV) to follow natural-language instructions and navigate through complex 3D aerial environments. Recent advances in multimodal large language models (MLLMs) have inspired a series of zero-shot methods for Aerial VLN~\cite{chen2026vision, wang2025towards}. Compared with learning-based approaches, these methods require neither task-specific navigation training nor large-scale annotated trajectories, offering a promising path toward more generalizable UAV navigation. However, generic MLLMs remain limited in their ability to infer precise metric and directional relations directly from visual observations~\cite{zhang2025open3d}. Zero-shot Aerial VLN therefore requires explicit and interpretable spatial representations to ground model reasoning in environmental geometry and support accurate navigation decisions~\cite{xia2026vision}.

\begin{figure}[t]
  \includegraphics[width=\linewidth]{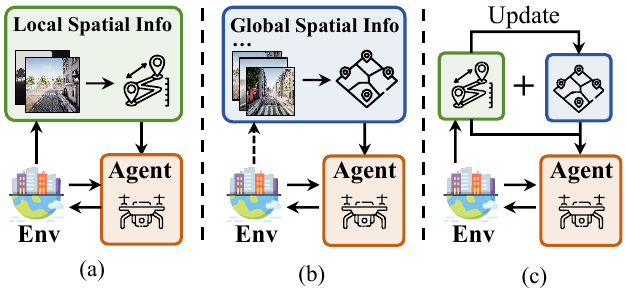}
  \caption{Comparison of three spatial representation paradigms in zero-shot methods: (a) the local-centric paradigm, (b) the global-centric paradigm, and (c) our anchor-centric paradigm, which bridges spatial information across both scales.}
  \label{fig:gap}
\end{figure}

As illustrated in Figure~\ref{fig:gap}, existing zero-shot methods typically represent spatial information on either a local or global scale. Local-centric methods construct structured spatial representations online from visual observations, such as 2D semantic maps~\cite{gao2024exploring} or depth-aware geometric cues~\cite{hu2025see, zheng2026onfly}. In contrast, global-centric methods organize historical experience into large-scale memory graphs~\cite{zhang2025citynavagent} or landmark knowledge bases~\cite{ning2026lookasidevln}, which are accessed through graph search or knowledge retrieval. However, relying on either scale alone is inherently limiting. Local spatial information provides the geometric evidence required to ground immediate actions in the surrounding environment, whereas global spatial information provides scene-level priors that align long-horizon path planning with instructions, particularly when instruction-relevant landmarks lie beyond the UAV's current field of view. A straightforward way to combine their strengths is to model the two scales independently and fuse their representations during decision-making. However, such separation overlooks their inherent compositional relationship: locally grounded spatial evidence can naturally evolve into persistent global knowledge. Maintaining the two scales separately makes such cross-scale accumulation indirect and requires the decision model to reconcile heterogeneous spatial representations at decision time, complicating their effective integration. These limitations motivate us to design a shared spatial primitive that retains the complementary benefits of local and global spatial representations while enabling locally grounded evidence to be progressively accumulated into global memory and allowing both scales to be jointly exploited for navigation.

To this end, we introduce AirAnchor, a new paradigm that bridges local spatial grounding and global spatial memory within an MLLM-powered navigation agent through structured prompts. Our key insight is that Aerial VLN does not require an exhaustive spatial representation of the environment. Instead, decision-relevant spatial information can be distilled into a sparse set of semantically meaningful references, which we term spatial anchors. We define two types of anchors: object anchors, which represent specific objects and their spatial attributes, and directional anchors, which encode spatial cues associated with particular directions. Anchors constructed from real-time observations can be organized into local spatial representations, while object anchors can be persistently accumulated to form global spatial representations. To realize this idea, we design a UAV navigation system with a modular architecture comprising three main components: (1) \textbf{Query-Driven Spatial Anchor Grounding} queries the MLLM to identify decision-relevant object and directional anchors in the current observation, computes their spatial cues from the corresponding depth map, and organizes them into an Egocentric Anchor Graph (EAG) as the local spatial representation. (2) \textbf{Persistent Object Spatial Memory} maintains an object knowledge base as the global spatial representation for retrieving landmark priors. During navigation, landmarks extracted from the instruction are used as queries to retrieve candidate objects from the knowledge base. The MLLM then selects the candidate that best matches the referenced landmark. Meanwhile, newly acquired object anchors are either merged with existing instances or inserted as new entries through similarity-based matching. (3) A \textbf{Spatially-Informed Navigation Agent} is an MLLM-based agentic framework that integrates subtask management, skill selection, and progress reflection into a closed-loop decision-making pipeline. It jointly aligns the UAV's current observation and user instruction with the local and global spatial information provided by the EAG and landmark priors, enabling comprehensive spatial grounding for navigation. Extensive experiments on AerialVLN~\cite{liu2023aerialvln} demonstrate that AirAnchor consistently improves the performance of zero-shot agents, validating the effectiveness of our paradigm.

The contributions of our method are summarized as follows: (1) We introduce AirAnchor, a novel paradigm that bridges local spatial grounding and global spatial memory through spatial anchors, allowing information from both scales to be jointly utilized within a navigation framework. (2) To implement AirAnchor, we develop a modular system consisting of three components: Query-Driven Spatial Anchor Grounding, Persistent Object Spatial Memory, and the Spatially-Informed Navigation Agent. (3) Extensive experiments demonstrate that AirAnchor outperforms existing zero-shot baselines. Further ablation studies validate the effectiveness of designed components.

\section{Related Works}

\noindent \textbf{Vision-and-Language Navigation.} Vision-and-Language Navigation (VLN), introduced with the Room-to-Room (R2R) benchmark~\cite{anderson2018vision}, was later extended to continuous environments with low-level control by R2R-CE~\cite{krantz2020beyond}. Early methods focused on cross-modal representation learning~\cite{hao2020towards,hong2021vln} and history modeling~\cite{chen2021history}, while recent advances in MLLMs have enabled vision-language-action models~\cite{wang2026limits}, fast--slow dual-system architectures~\cite{wei2026ground}, and zero-shot agentic frameworks~\cite{zhou2024navgpt,chen2024mapgpt}.

Aerial VLN~\cite{liu2023aerialvln} further extends VLN to continuous 3D aerial environments. Learning-based methods use trajectory supervision, with improvements from temporal history modeling~\cite{gao2025openfly}, map-aware action prediction~\cite{zhao2025aerial}, and multi-task learning~\cite{xu2026aerial}. Zero-shot methods instead exploit MLLM reasoning through structured local spatial representations~\cite{gao2024exploring,zheng2026onfly,hu2025see}, global memory~\cite{zhang2025citynavagent,ning2026lookasidevln}, and fine-grained modular design~\cite{shao2026finecog}. AirAnchor follows the zero-shot paradigm but differs from prior methods by bridging local spatial grounding and global spatial memory through spatial anchors and jointly exploiting both within an agentic framework.

\noindent \textbf{Spatial Representations for VLN.} Spatial representations ground navigation decisions in the environment. Early VLN methods implicitly encode spatial context through recurrent states or multimodal histories~\cite{anderson2018vision,chen2021history}. Later approaches introduce explicit local geometry using egocentric or top-down semantic maps~\cite{georgakis2022cross,gao2024exploring} and bird's-eye-view structures~\cite{liu2023bird} for action grounding. Global representations further accumulate navigation experience for long-horizon reasoning, typically through topological graphs~\cite{chen2021topological,chen2022think,zhang2025citynavagent} or compact landmark memories~\cite{ning2026lookasidevln}. However, local and global information is usually represented and maintained separately, making cross-scale information exchange and accumulation indirect. AirAnchor instead adopts object anchors as shared primitives across both scales, allowing spatial evidence grounded online to support immediate decisions while being naturally accumulated as persistent global knowledge.

\section{Method}
\begin{figure*}[t]
  \includegraphics[width=\linewidth]{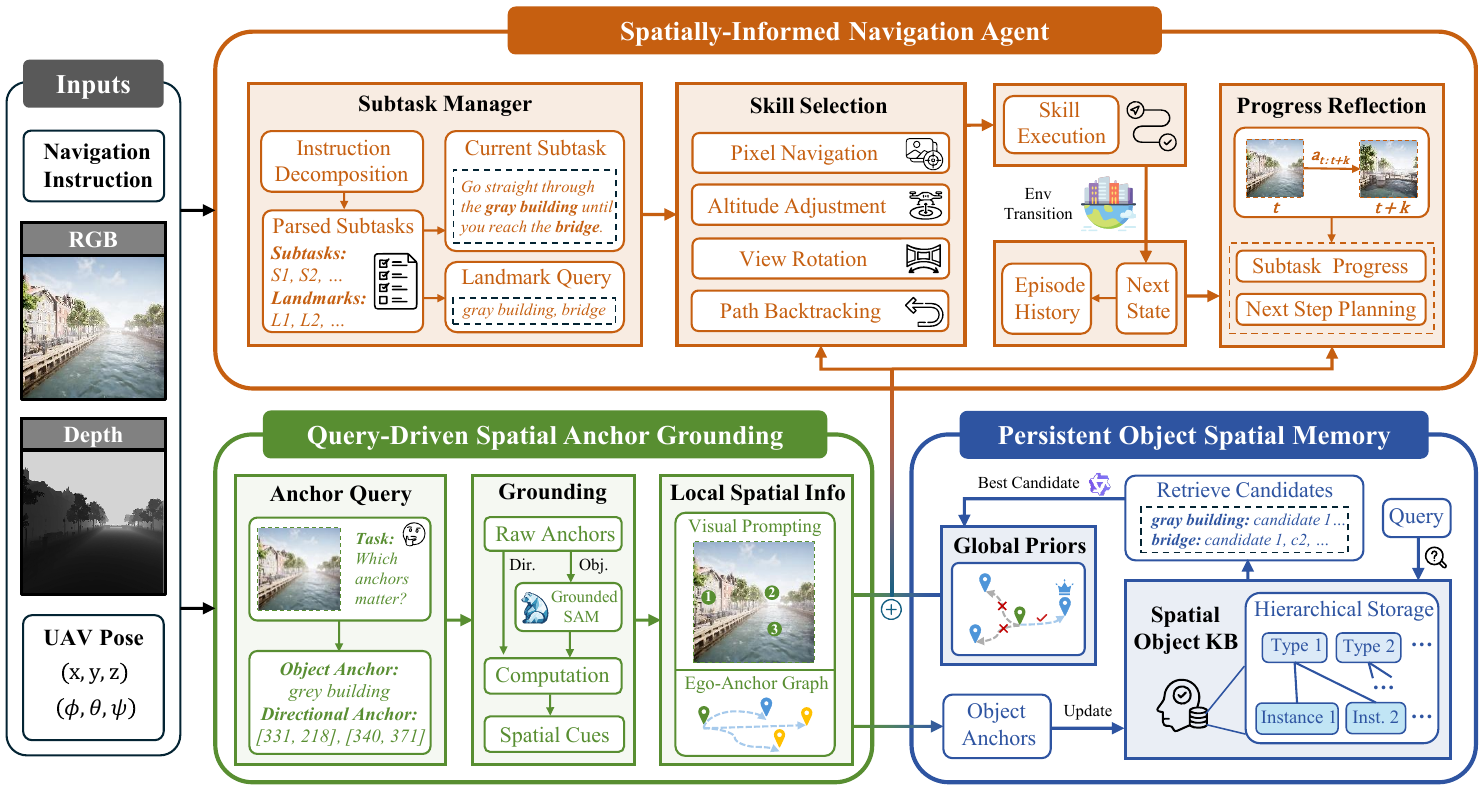}
  \caption{AirAnchor consists of three key modules. Query-Driven Spatial Anchor Grounding identifies spatial anchors and constructs the local spatial representation. Persistent Object Spatial Memory maintains the global spatial representation, supporting online updates and prior knowledge retrieval. Spatially-Informed Navigation Agent integrates the local spatial representation and global spatial priors within an agentic framework to enable closed-loop and spatially grounded navigation decisions.}
  \label{fig:framework}
\end{figure*}

In this section, we introduce the AirAnchor paradigm for zero-shot Aerial VLN. As illustrated in Figure~\ref{fig:framework}, at each decision step, \textbf{Query-Driven Spatial Anchor Grounding} identifies decision-relevant anchors from the current observation and organizes their spatial cues into an Egocentric Anchor Graph (EAG) as the local spatial representation. \textbf{Persistent Object Spatial Memory} consolidates observed object anchors into a Spatial Object Knowledge Base (SOKB) and retrieves instruction-relevant landmarks as global spatial priors. Finally, the \textbf{Spatially-Informed Navigation Agent} jointly exploits the local EAG and global priors within an MLLM-powered agentic framework for skill-level planning and progress reflection.

\subsection{Task Formulation}
\label{sec:task}

In Aerial VLN, a UAV agent receives a natural-language instruction $\mathcal{I}$ and navigates in a 3D urban environment. At timestep $t$, it observes an egocentric RGB-D observation $(I_t, D_t)$ and its pose $\mathbf{p}_t$. The UAV can execute eight discrete actions: \textit{Move Forward}, \textit{Turn Left}, \textit{Turn Right}, \textit{Ascend}, \textit{Descend}, \textit{Move Left}, \textit{Move Right}, and \textit{Stop}~\cite{liu2023aerialvln}. Navigation is successful if the UAV stops within a predefined distance threshold of the target destination. In our paradigm, rather than invoking the MLLM for every low-level action decision, AirAnchor performs high-level skill selection and delegates skill execution to a motion planner.

\subsection{Query-Driven Spatial Anchor Grounding}
\label{sec:anchor}

Aerial environments are large, open, and inherently 3D. Consequently, dense geometric maps are expensive to construct and maintain, whereas 2D grid maps or graphs may discard altitude information and fine-grained spatial relations. Our key insight is that, given the semantic reasoning capabilities of MLLMs, grounding the local spatial context does not require an exhaustive geometric representation. Instead, it is sufficient to expose the geometry of a sparse set of scene elements that are relevant to the current decision. We term these semantic-geometric references \textbf{spatial anchors}.

\noindent\textbf{Query-Driven Anchor Selection.} Let $\mathcal{C}_t=(s_n,\gamma_t,\pi_t)$ denote the navigation context, where $s_n$, $\gamma_t$, and $\pi_t$ represent the current subtask, navigation progress, and plan, respectively. Conditioned on $I_t$ and $\mathcal{C}_t$, the MLLM queries
\begin{equation}
    \mathcal{A}_t
    =
    \mathrm{MLLM}_{\mathrm{query}}(I_t,\mathcal{C}_t)
    =
    \mathcal{A}_t^{o}\cup\mathcal{A}_t^{d},
    \label{eq:anchor_query}
\end{equation}
where $\mathcal{A}_t^{o}$ and $\mathcal{A}_t^{d}$ denote object anchors and directional anchors, respectively. Object anchors identify important objects through semantic labels, whereas directional anchors specify ray-casting directions through pixel coordinates. This query-driven design enables the MLLM to actively identify key elements in $I_t$ whose spatial information is most relevant to the current navigation decision.

\noindent\textbf{Spatial Cue Grounding.} For each queried anchor $a_i$, we derive its spatial cues from the depth $D_t$. Specifically, for a directional anchor, we back-project its pixel coordinate and depth value to obtain the corresponding reference point in the world frame. For an object anchor, an open-vocabulary detector based on Grounded SAM~\cite{ren2024grounded} first localizes and segments the object, after which the corresponding depth pixels are back-projected to obtain its 3D point cloud. To compactly characterize its geometry, we represent the object using a 2.5D spatial extent $\mathbf{g}_i=(R_i,z_i^{\min},z_i^{\max})$, where $R_i$ denotes its horizontal footprint and $[z_i^{\min},z_i^{\max}]$ denotes its vertical extent. The object center $\mathbf{c}_i$ is computed from $\mathbf{g}_i$ and used as its reference point. We then characterize the spatial cues of each anchor $a_i$ by computing the relative geometry between its reference point and the UAV pose:
\begin{equation}
    \mathbf{r}_{t,i}
    =
    (\phi_{t,i},
    \Delta z_{t,i},
    d^{xy}_{t,i},
    d^{3D}_{t,i}),
    \label{eq:spatial_cue}
\end{equation}
where $\phi_{t,i}$ denotes the relative bearing, $\Delta z_{t,i}$ the relative height difference, and $d^{xy}_{t,i}$ and $d^{3D}_{t,i}$ the horizontal and Euclidean distances, respectively.

\noindent\textbf{Egocentric Anchor Graph.} We organize the grounded anchors into an Egocentric Anchor Graph (EAG), denoted by $\mathcal{G}_t=(\mathcal{V}_t,\mathcal{E}_t)$, where the node set $\mathcal{V}_t$ and edge set $\mathcal{E}_t$ are defined as
\begin{equation}
\begin{aligned}
\mathcal{V}_t &= \{v_t^{u}\} \cup \{v_i \mid a_i \in \mathcal{A}_t\},\\
\mathcal{E}_t &= \{(v_t^{u}, v_i) \mid a_i \in \mathcal{A}_t\}.
\end{aligned}
\end{equation}
Here, $v_t^{u}$ and $v_i$ denote the UAV node and anchor node, respectively. Each anchor node stores its semantic and geometric attributes, while each UAV-anchor edge encodes the corresponding spatial cue $\mathbf{r}_{t,i}$. In addition, we annotate the anchors in $I_t$ with visual prompts to obtain $I'_t$, explicitly aligning the structured EAG with the visual observation.

Through the design of spatial anchors, we construct a lightweight and interpretable local spatial representation that provides explicit local grounding for MLLM-based navigation decisions. Implementation details are provided in the Appendix~\ref{app:anchor}.

\begin{figure}[t]
  \includegraphics[width=\linewidth]{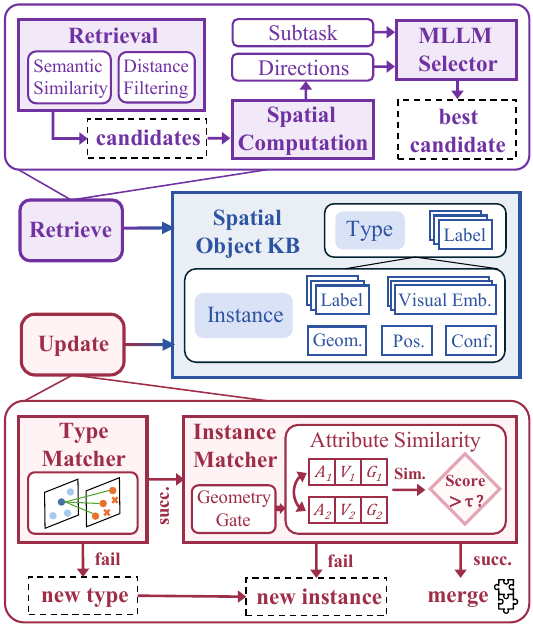}
  \caption{Overview of Persistent Object Spatial Memory. Centered on the SOKB, it maintains the global spatial representation to support reliable landmark-prior retrieval and online object knowledge updates.}
  \label{fig:memory}
\end{figure}

\subsection{Persistent Object Spatial Memory}
\label{sec:memory}

Long-horizon Aerial VLN often refers to landmarks beyond the UAV's current field of view, requiring global spatial representations to guide path planning. In this setting, an object-centric knowledge base provides an efficient representation for large urban scenes. However, urban scenes exhibit complex and diverse object semantics and visual appearances, making robust object association and memory maintenance challenging. Therefore, we design our memory around a Spatial Object Knowledge Base (SOKB), which adopts a hierarchical storage structure together with coarse-to-fine object association and knowledge retrieval mechanisms.

\noindent\textbf{Hierarchical Structure.} As shown in Figure~\ref{fig:memory}, the SOKB $\mathcal{M}$ adopts a type-instance hierarchy:
\begin{equation}
\begin{aligned}
    \mathcal{M}
    &= \{\mathcal{T}_m\}_{m=1}^{N_T},
    \;
    \mathcal{T}_m
    = (\mathcal{B}^{T}_m,\mathcal{O}_m), \\
    \mathcal{O}_m
    &= \{o_j\}_{j=1}^{N_m},
    \;
    o_j
    =
    (\mathcal{B}^{A}_j,
     \mathcal{B}^{V}_j,
     \mathbf{g}_j,
     \mathbf{c}_j,
     q_j).
\end{aligned}
\label{eq:memory}
\end{equation}
Here, each type $\mathcal{T}_m$ represents an object category and $\mathcal{O}_m$ contains the distinct object instances belonging to that category. We employ bounded banks to store category labels $\mathcal{B}^{T}_m$, appearance labels $\mathcal{B}^{A}_j$, and visual embeddings $\mathcal{B}^{V}_j$, respectively mitigating synonymous category expressions, diverse object appearances, and substantial viewpoint-dependent visual variations in Aerial VLN. For each instance $o_j$, $\mathbf{g}_j$ denotes its geometric extent, $\mathbf{c}_j$ its 3D center, and $q_j$ its confidence score. This hierarchy supports efficient organization and coarse-to-fine association and retrieval.

\noindent\textbf{Object Association.} A newly observed object anchor $a_i^o$ is first matched against object types according to semantic similarity. Candidate instances under compatible types are then filtered by a geometry gate to remove spatially implausible matches. For each remaining instance $o_j$, we compute a weighted association score:
\begin{equation}
\begin{aligned}
S_M(i,j)
&= w_G S_G(i,j) + w_S S_S(i,j) \\
&\quad + w_V S_V(i,j),
\end{aligned}
\label{eq:matching}
\end{equation}
where $S_G$ measures geometric consistency between spatial extents, $S_S$ captures category- and appearance-level semantic similarity, and $S_V$ measures visual similarity using DINOv2~\cite{oquab2024tmlr-dinov2} embeddings. If the highest association score exceeds a predefined threshold, the incoming anchor is fused with the matched instance by merging their spatial extents and updating the semantic and visual banks; otherwise, a new instance is created.

\noindent\textbf{Landmark Prior Retrieval.}
For each active subtask $s_n$, its referenced landmarks $\mathcal{L}_n$ are used to retrieve candidate instances from the SOKB based on semantic similarity $S_S$ and a distance filter, yielding a candidate set $\mathcal{C}_{\ell}$ for each landmark $\ell\in\mathcal{L}_n$. We define
$\mathcal{R}_{t,\ell}=\{(o_j,\mathbf{r}_{t,j})\mid o_j\in\mathcal{C}_{\ell}\}$
as the candidate instances together with their UAV-relative spatial cues. The MLLM is then employed for candidate selection:
\begin{equation}
\hat{\mathcal{O}}_n
=
\mathrm{MLLM}_{\mathrm{ret}}
\big(
s_n,
{\{(\ell,\mathcal{R}_{t,\ell})}\mid{\ell\in\mathcal{L}_n\}}
\big),
\label{eq}
\end{equation}
where $\hat{\mathcal{O}}_n=\{\hat{o}_{\ell}\mid\ell\in\mathcal{L}_n\}$ contains the best candidate $\hat{o}_{\ell}$ selected for each landmark. The MLLM disambiguates semantically similar candidates by matching their spatial cues with the spatial relations implied by the instruction to identify the referred landmark. The selected instances and their corresponding spatial cues are organized into the global landmark prior
$\mathcal{P}_t=\{(\hat{o}_{\ell},\mathbf{r}_{t,\hat{o}_{\ell}})\mid\ell\in\mathcal{L}_n\}$.

Overall, our memory design enables efficient maintenance and online updates while providing reliable landmark priors for long-horizon navigation. Detailed object association, merging, and retrieval procedures are provided in Appendix~\ref{app:memory}.

\subsection{Spatially-Informed Navigation Agent}
\label{sec:agent}

To integrate spatial information across local and global scales for robust and spatially grounded navigation, we introduce a Spatially-Informed Navigation Agent that combines subtask management, skill-level planning, and progress reflection in a closed-loop framework, while seamlessly incorporating spatial representations at both scales through structured prompts.

\noindent\textbf{Subtask Management.} At the beginning of each episode, the instruction is decomposed into an ordered sequence of subtasks:
\begin{equation}
    \{(s_n,\mathcal{L}_n)\}_{n=1}^{N}
    =
    \mathrm{MLLM}_{\mathrm{dec}}(\mathcal{I}),
    \label{eq:subtask}
\end{equation}
where $s_n$ denotes a subtask and $\mathcal{L}_n$ its referenced landmarks. This allows the agent to focus on one subtask at a time, reducing the burden of long-horizon instruction tracking.

\noindent\textbf{Skill-Level Planning.}
We employ the MLLM as a high-level planner that performs skill-level decision making. Let
$\sigma_t=k_t(\boldsymbol{\eta}_t)$
denote a parameterized skill, where $k_t$ specifies the skill type and $\boldsymbol{\eta}_t$ its parameters. The planning process is formulated as
\begin{equation}
    (\sigma_t,\xi_t)
    =
    \mathrm{MLLM}_{\mathrm{nav}}
    \left(
    I'_t,\mathcal{C}_t,
    \mathcal{G}_t,\mathcal{P}_t
    \right),
    \label{eq:navigation}
\end{equation}
where $\xi_t$ denotes the rationale for the selected skill.

We define four skills tailored to Aerial VLN: \textit{Pixel Navigation} specifies a target waypoint using a designated image coordinate and depth value; \textit{Altitude Adjustment} controls vertical displacement; \textit{View Rotation} determines a turning angle from panoramic observations; and \textit{Path Backtracking} recovers from navigation errors by returning to a previous waypoint. The selected skill $\sigma_t$ is converted into a target pose and executed by a low-level motion planner. Further details of skill-level planning are provided in Appendix~\ref{app:skill}.

\noindent\textbf{Progress Reflection.}
Reliable reasoning about navigation progress often depends on changes in spatial relations rather than solely on textual or visual semantics, such as when determining whether the UAV has passed or flown over a landmark. We therefore explicitly incorporate spatial context into the reflection process. Specifically, after executing skill $\sigma_t$ for $k$ low-level actions, the UAV moves from $\mathbf{p}_t$ to $\mathbf{p}_{t+k}$. We retain the anchors in $\mathcal{G}_t$ and recompute their spatial cues relative to $\mathbf{p}_{t+k}$, yielding $\widetilde{\mathcal{G}}_{t+k}$, which encodes the spatial relations between the current pose and the anchors in $\mathcal{G}_t$. We then summarize the reflection inputs into a semantic context
$\mathcal{C}^{\mathrm{sem}}_t=(\mathcal{C}_t,I'_t,I'_{t+k},\sigma_t,\xi_t)$
and a spatial context
$\mathcal{C}^{\mathrm{spa}}_t=(\mathcal{G}_t,\widetilde{\mathcal{G}}_{t+k})$.
The reflection process is formulated as
\begin{equation}
    (\gamma_{t+k},b_{t+k},\pi_{t+k})
    =
    \mathrm{MLLM}_{\mathrm{ref}}
    (\mathcal{C}^{\mathrm{sem}}_t,
     \mathcal{C}^{\mathrm{spa}}_t),
    \label{eq:reflection}
\end{equation}
where $\gamma_{t+k}$ denotes the updated assessment of navigation progress, $b_{t+k}$ is a binary variable indicating whether the current subtask has been completed, and $\pi_{t+k}$ specifies the plan for the next step.

Overall, the agent integrates local and global spatial information into an agentic framework, enabling efficient and spatially grounded navigation.

\section{Experiments}
\subsection{Experimental Setup}
\noindent\textbf{Dataset.} We evaluate AirAnchor on the AerialVLN-S dataset of the challenging AerialVLN~\cite{liu2023aerialvln} benchmark. The benchmark comprises 8,446 flight trajectories collected from experienced UAV pilots across 25 diverse city-scale environments built in Unreal Engine 4, covering more than 870 categories of urban objects. AerialVLN-S is a small-scale variant of AerialVLN consisting of 17 compact scenes.

\noindent\textbf{Metrics.} Following AerialVLN, we report four evaluation metrics: Success Rate (SR), Oracle Success Rate (OSR), Navigation Error (NE), and Success weighted by normalized Dynamic Time Warping (SDTW). SR measures the percentage of episodes in which the agent stops within 20 meters of the target, whereas OSR considers an episode successful if any point along the predicted trajectory enters the 20\,m success radius. NE is the Euclidean distance between the agent's final position and the target. SDTW jointly measures navigation success and trajectory fidelity with respect to the reference path.

\noindent\textbf{Baselines.}
We compare AirAnchor with three groups of baselines. \textit{Statistical methods} include Random and Action Sampling. \textit{Learning-based methods} include Seq2Seq~\cite{anderson2018vision}, CMA~\cite{krantz2020beyond}, and LAG~\cite{liu2023aerialvln}, which learn navigation policies from task-specific trajectory supervision. For \textit{zero-shot methods}, we include indoor VLN agents NavGPT~\cite{zhou2024navgpt} and MapGPT~\cite{chen2024mapgpt}, as well as Aerial VLN agents TypeFly~\cite{chen2023typefly}, PIVOT~\cite{nasiriany2024pivot}, SPF~\cite{hu2025see}, and FineCog-Nav~\cite{shao2026finecog}.

\noindent\textbf{Implementation Details.}
We implement AirAnchor with Qwen3.6-Plus~\cite{qwen36plus} as the MLLM backbone via its online API. A$^\ast$ search~\cite{hart1968formal} serves as the low-level motion planner, generating feasible paths from the current pose to the target pose. Each scene starts with an empty SOKB, which is updated online and retained across episodes within the same scene. All zero-shot Aerial VLN baselines are reproduced using the same MLLM backbone.

\begin{table*}[t]
\centering
\small
\renewcommand{\arraystretch}{1.12}
\resizebox{\linewidth}{!}{%
\begin{tabular}{
c
c
c
*{8}{c}
}
\toprule
\multicolumn{2}{c}{\multirow[c]{2}{*}{Category}}
& \multirow[c]{2}{*}{Method}
& \multicolumn{4}{c}{Validation Seen}
& \multicolumn{4}{c}{Validation Unseen}
\\
\cmidrule(lr){4-7}
\cmidrule(lr){8-11}
\multicolumn{2}{c}{}
&
& SR$\uparrow$
& OSR$\uparrow$
& SDTW$\uparrow$
& NE$\downarrow$
& SR$\uparrow$
& OSR$\uparrow$
& SDTW$\uparrow$
& NE$\downarrow$
\\
\midrule

\multicolumn{2}{c}{
\multirow[c]{2}{*}{Statistical}
}
& Random
& 0.0
& 0.0
& 0.0
& 109.6
& 0.0
& 0.0
& 0.0
& 149.7
\\
\multicolumn{2}{c}{}
& Action Sampling
& 0.9
& 5.7
& 0.3
& 213.8
& 0.2
& 1.1
& 0.1
& 237.6
\\

\midrule

\multicolumn{2}{c}{
\multirow[c]{3}{*}{Learning-Based}
}
& Seq2Seq
& 4.8
& 19.8
& 1.6
& 146.0
& 2.3
& 11.7
& 0.7
& 218.9
\\
\multicolumn{2}{c}{}
& CMA
& 3.0
& \textbf{23.2}
& 0.6
& 121.0
& 3.2
& \underline{16.0}
& 1.1
& 172.1
\\
\multicolumn{2}{c}{}
& LAG
& \underline{7.2}
& 15.7
& 2.4
& \underline{90.2}
& 5.1
& 10.5
& 1.4
& 127.9
\\

\midrule

\multirow[c]{7}{*}{Zero-Shot}
& \multirow[c]{2}{*}{Indoor}
& NavGPT
& 0.0
& 0.0
& 0.0
& 163.5
& 0.0
& 0.0
& 0.0
& \textbf{82.1}
\\
&
& MapGPT
& 2.1
& 4.7
& 0.8
& 124.9
& 0.0
& 0.0
& 0.0
& 107.0
\\

\cmidrule(lr){2-11}

& \multirow[c]{5}{*}{Aerial}
& TypeFly
& 1.8
& 5.5
& 0.4
& 136.8
& 1.4
& 4.3
& 0.3
& 166.4
\\
&
& PIVOT
& 3.2
& 12.4
& 0.9
& 116.3
& 2.6
& 10.2
& 0.7
& 139.6
\\
&
& SPF
& 5.7
& 14.4
& 2.1
& 103.8
& 5.1
& 12.2
& 1.7
& 116.0
\\
&
& FineCog-Nav
& 6.9
& 16.8
& \underline{2.7}
& 97.5
& \underline{6.2}
& 15.3
& \underline{2.3}
& 107.0
\\
&
& \textbf{AirAnchor (Ours)}
& \textbf{9.6}
& \underline{23.1}
& \textbf{4.0}
& \textbf{80.5}
& \textbf{9.2}
& \textbf{22.0}
& \textbf{3.5}
& \underline{84.6}
\\

\bottomrule
\end{tabular}%
}

\caption{Overall performance comparison on the AerialVLN-S benchmark. 
Bold and underline indicate the best and second-best results, respectively.}
\label{tab:main_results}
\end{table*}

\begin{table}[t]
\centering
\footnotesize
\setlength{\tabcolsep}{3.2pt}
\renewcommand{\arraystretch}{1.06}
\begin{tabular}{clccc}
\toprule
Module & Ablation Variant
& SR$\uparrow$
& SDTW$\uparrow$
& NE$\downarrow$ \\
\midrule
-- & \textbf{AirAnchor}
& \textbf{9.6} & \textbf{4.0} & \textbf{80.5} \\
\midrule
\multirow[c]{2}{*}{Anchor}
& w/o Dir. Anchor & 9.3 & 3.7 & 84.7 \\
& w/o Obj. Anchor & 8.7 & 3.5 & 87.6 \\
\midrule
\multirow[c]{2}{*}{Memory}
& Flat KB & 8.4 & 3.4 & 90.8 \\
& w/o MLLM Selector & 9.0 & 3.6 & 86.9 \\
\midrule
\multirow[c]{5}{*}{Agent}
& w/o Subtask & 8.7 & 3.5 & 88.6 \\
& Action-Level Planning & 9.0 & 3.3 & 85.6 \\
& w/o EAG & 7.8 & 2.9 & 93.8 \\
& w/o Landmark Prior & 8.1 & 3.2 & 94.5 \\
& w/o Spatial Reflection & 8.4 & 3.3 & 89.0 \\
\bottomrule
\end{tabular}
\caption{Ablation study of different modules.}
\label{tab:ablation}
\end{table}

\begin{figure}[ht]
  \includegraphics[width=\linewidth]{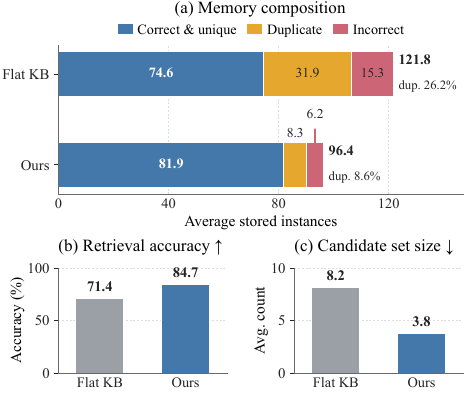}
    \caption{
    Comparison of object knowledge base design.
    We compare the proposed SOKB with a flat KB in terms of
    (a) memory composition, (b) landmark retrieval accuracy, and
    (c) candidate-set size.
    }
  \label{fig:mem_aba}
\end{figure}

\begin{table}[ht]
\centering
\footnotesize
\setlength{\tabcolsep}{2.7pt}
\renewcommand{\arraystretch}{1.08}
\begin{tabular}{lccccccc}
\toprule
Setting & Pixel & Alt. & View & Back.
& SR$\uparrow$ & SDTW$\uparrow$ & NE$\downarrow$ \\
\midrule
\textbf{Full}
& $\checkmark$ & $\checkmark$ & $\checkmark$ & $\checkmark$
& \textbf{9.6} & \textbf{4.0} & \textbf{80.5} \\
$-$ Alt.
& $\checkmark$ & $\times$ & $\checkmark$ & $\checkmark$
& 9.0 & 3.6 & 85.8 \\
$-$ View
& $\checkmark$ & $\checkmark$ & $\times$ & $\checkmark$
& 8.6 & 3.4 & 87.8 \\
$-$ Back.
& $\checkmark$ & $\checkmark$ & $\checkmark$ & $\times$
& 9.3 & 3.8 & 83.1 \\
Pixel Only
& $\checkmark$ & $\times$ & $\times$ & $\times$
& 8.1 & 3.1 & 91.6 \\
\bottomrule
\end{tabular}
\caption{Ablation of the navigation skill set.}
\label{tab:skill_ablation}
\end{table}

\subsection{Overall Performance}
\label{sec:overall}

Table~\ref{tab:main_results} summarizes the overall results on AerialVLN-S. AirAnchor performs strongly on both validation splits and outperforms all learning-based baselines on Validation Unseen, demonstrating robust zero-shot generalization across aerial environments.
In contrast, indoor zero-shot agents such as NavGPT and MapGPT achieve nearly zero success when directly transferred to aerial environments. This suggests that ground-based zero-shot VLN methods do not transfer effectively to Aerial VLN, where dedicated spatial reasoning mechanisms are required to handle large-scale 3D aerial environments.

Among zero-shot aerial methods, AirAnchor consistently outperforms all baselines across both validation splits.
Compared with the strongest zero-shot aerial baseline, FineCog-Nav, AirAnchor achieves relative improvements of 39.1\% / 48.4\% in SR, 37.5\% / 43.8\% in OSR, and 48.1\% / 52.2\% in SDTW on Seen / Unseen, while reducing NE by 17.4\% / 20.9\%, respectively.
The substantial improvements in OSR and NE are consistent with the complementary use of local metric grounding and persistent landmark priors, which provides more reliable spatial guidance toward instruction-relevant regions, including those beyond the current field of view.
Meanwhile, the gains in SR and SDTW suggest that spatially informed skill-level planning and progress reflection more effectively translate spatial knowledge into successful and trajectory-aligned navigation.
Notably, AirAnchor exhibits minor degradation from Validation Seen to Unseen, highlighting its robustness across diverse aerial scenes.

\begin{figure*}[t]
  \includegraphics[width=\linewidth]{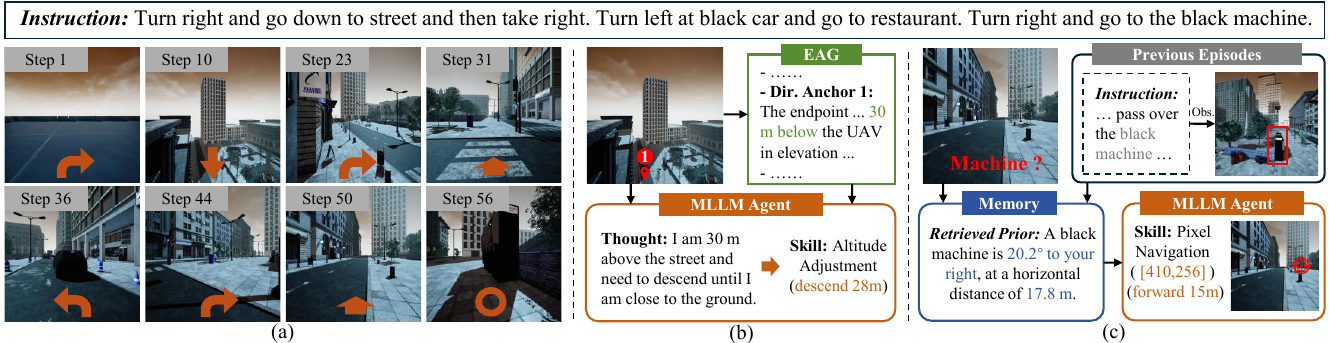}
  \caption{Visualization of a successful navigation episode. (a) Key steps throughout the episode. (b) Local spatial grounding at Step 10. (c) Global prior guidance at Step 50.}
  \label{fig:case}
\end{figure*}

\subsection{Ablation Study}
\label{sec:ablation}

To comprehensively evaluate AirAnchor, we conduct ablation studies on its key components. All ablation experiments are performed on the validation "Seen" split of the AerialVLN-S benchmark.

\noindent\textbf{Effect of Spatial Anchors.}
As shown in Table~\ref{tab:ablation}, removing directional anchors reduces SR by 3.1\%, while removing object anchors causes larger drops in SR and SDTW (9.4\% and 12.5\%) and increases NE by 7.1\,m. This confirms that directional anchors provide useful ray-based geometry, while object anchors additionally ground instruction-relevant landmarks with semantic and metric cues.

\noindent\textbf{Effect of Persistent Object Memory.}
Replacing SOKB with \textit{Flat KB}, which indexes objects by labels and merges same-label observations using a fixed spatial threshold~\cite{ning2026lookasidevln}, reduces SR by 12.5\% and increases NE by 10.3\,m. Replacing the MLLM selector with semantic-nearest retrieval results in a smaller degradation, confirming the benefit of spatially informed landmark disambiguation. Figure~\ref{fig:mem_aba} further shows that SOKB stores 20.9\% fewer instances while preserving 9.8\% more correct unique objects, reduces duplicate instances by 74.0\%, and improves retrieval accuracy by 18.6\%. It also reduces the candidate-set size by 53.7\%, indicating more efficient retrieval over the object knowledge. These results demonstrate more reliable and compact spatial memory.

\noindent\textbf{Effect of Agent Design.}
Table~\ref{tab:ablation} shows that removing subtask management degrades long-horizon navigation, while action-level planning reduces SR by 6.3\% and SDTW by 17.5\%, supporting skill-level decision making. Removing EAG yields the largest SR and SDTW drops (18.8\% and 27.5\%), whereas removing landmark priors causes the largest NE increase (14.0\,m), highlighting the complementary roles of local geometry and global spatial guidance. Removing spatial context from reflection further reduces SR and SDTW by 12.5\% and 17.5\%, respectively, confirming the benefit of explicit spatial context for progress reasoning. Finally, Table~\ref{tab:skill_ablation} shows that all specialized skills are beneficial, with \textit{View Rotation} having the largest impact, followed by \textit{Altitude Adjustment} and \textit{Path Backtracking}; using \textit{Pixel Navigation} alone reduces SR by 15.6\% and increases NE by 11.1\,m.

\subsection{Case Analysis}
\label{sec:case}

Figure~\ref{fig:case} illustrates how AirAnchor coordinates local and global spatial information within a long-horizon episode.
At Step~10, the EAG grounds a directional anchor whose endpoint is about 30\,m below the UAV.
The agent uses this metric cue to select \textit{Altitude Adjustment} and descend 28\,m toward street level, showing how local anchors support precise 3D control.
At Step~50, when the instruction refers to the black machine, the SOKB retrieves a previously observed instance and provides a prior locating it $20.2^\circ$ to the right and 17.8\,m away.
The agent then selects \textit{Pixel Navigation} toward the landmark.
This example illustrates the complementary roles of local metric grounding and persistent global guidance during navigation.

\section{Conclusion}
\label{sec:conclusion}

In this work, we introduced AirAnchor, a zero-shot Aerial VLN paradigm that bridges local grounding and global memory through spatial anchors within an MLLM-powered agentic framework. AirAnchor grounds spatial anchors into an Egocentric Anchor Graph for local spatial reasoning, while persistently consolidating object anchors into a Spatial Object Knowledge Base to retrieve global spatial priors. A Spatially-Informed Navigation Agent further integrates both representations into a closed-loop process. The experimental results demonstrate the efficacy and robustness of our method.

\section{Limitations}

Despite its effectiveness, AirAnchor has several limitations. First, spatial anchors are grounded from RGB-D observations and UAV poses with open-vocabulary perception models. Errors in depth estimation, object localization, or geometric grounding can therefore propagate into the EAG and subsequently affect navigation decisions. For object anchors, incorrect association may further be accumulated in the persistent SOKB and influence later retrieval.

Second, the benefit of global spatial memory depends on previously accumulated scene knowledge. When entering a new environment, instruction-relevant landmarks may not yet exist in the SOKB, and the agent must rely primarily on local observations until sufficient object knowledge has been collected. Improving cold-start exploration and uncertainty-aware memory updates would make persistent spatial reasoning more robust.

Third, the current navigation agent employs a predefined set of four skills together with an A$^*$ low-level planner. Although this design provides an effective abstraction for AerialVLN-S, more complex maneuvers or dynamic environments may require adaptive or learned skill composition. Finally, our evaluation is limited to simulated AerialVLN-S environments, and the robustness of AirAnchor to real-world sensing noise and dynamic environments remains to be validated. Moreover, the current system relies on multiple foundation models, including an online MLLM, introducing additional computational and communication overhead. Future work should investigate more efficient implementations and the robustness of AirAnchor across different MLLM backbones.

\bibliography{custom}

\clearpage

\appendix

\section{Implementation Details}
\label{app:implementation}
This section provides additional implementation details of AirAnchor.

\subsection{Query-Driven Spatial Anchor Grounding}
\label{app:anchor}

\subsubsection{Anchor Query}
At decision step $t$, the anchor-query MLLM receives the current egocentric
RGB observation $I_t$ and navigation context
$\mathcal{C}_t=(s_n,\gamma_t,\pi_t)$. We query exactly $K=3$ spatial anchors,
\begin{equation}
    \mathcal{A}_t
    =
    \mathcal{A}_t^{o}
    \cup
    \mathcal{A}_t^{d},
\end{equation}
where the relative numbers of object and directional anchors are determined
by the MLLM according to the current navigation context. Object anchors
provide a category label $\ell_i^T$ and an appearance description
$\ell_i^A$, whereas directional anchors specify an image coordinate
$\mathbf{u}_i=(u_i,v_i)$. Directional anchors are transient local references
and are never inserted into the persistent SOKB. The query is explicitly
conditioned on the active navigation objective: the MLLM is instructed to
select complementary anchors whose geometry is useful for resolving the
current spatial decision rather than exhaustively describing visually salient
scene elements. Object-anchor bounding boxes, masks, and detector confidence
are not predicted by the MLLM; they are obtained by the grounding module
described below.

\subsubsection{Depth Back-Projection}
\label{app:depth_grounding}
Let $\mathbf{K}$ denote the camera intrinsic matrix and
$\mathbf{T}^{wc}_t=[\mathbf{R}^{wc}_t,\mathbf{t}^{wc}_t]$ the camera-to-world
transformation in this $z$-up frame. For a pixel $\mathbf{u}=(u,v)$ with
homogeneous coordinate $\widetilde{\mathbf{u}}=(u,v,1)^\top$, we first define
the unit camera ray
\begin{equation}
    \widehat{\mathbf{r}}^{c}(\mathbf{u})
    =
    \frac{\mathbf{K}^{-1}\widetilde{\mathbf{u}}}
    {\left\|\mathbf{K}^{-1}\widetilde{\mathbf{u}}\right\|_2}.
    \label{eq:app_unit_ray}
\end{equation}
For perspective depth $d(\mathbf{u})$, the corresponding world-coordinate
point is therefore
\begin{equation}
    \mathbf{q}^{w}(\mathbf{u},d)
    =
    \mathbf{R}^{wc}_t
    \left[
        d(\mathbf{u})\widehat{\mathbf{r}}^{c}(\mathbf{u})
    \right]
    +
    \mathbf{t}^{wc}_t.
    \label{eq:app_backprojection}
\end{equation}
The ray normalization in Eq.~\ref{eq:app_unit_ray} is required because the
depth value measures distance along the projection ray rather than planar
camera-axis depth.

We use a maximum reliable depth $D_{\max}=100~\mathrm{m}$. Depth values
beyond $D_{\max}$ are not interpreted as accurate metric surface
measurements. Instead, they indicate that the corresponding viewing ray
remains unobstructed beyond the reliable sensing range.

For a directional anchor $a_i^d$, we compute the median positive depth within
a $5\times5$ neighborhood $\mathcal{N}(\mathbf{u}_i)$:
\begin{equation}
    \bar d_i
    =
    \operatorname{median}
    \left\{
        D_t(\mathbf{u})
        \,\middle|\,
        \mathbf{u}\in\mathcal{N}(\mathbf{u}_i),
        D_t(\mathbf{u})>0
    \right\}.
    \label{eq:app_dir_depth}
\end{equation}
If no positive depth is available in the neighborhood, the anchor is
discarded. Otherwise, we define
\begin{equation}
    \widehat d_i
    =
    \min(\bar d_i,D_{\max}),
    \qquad
    f_i^{\mathrm{far}}
    =
    \mathbbm{1}[\bar d_i>D_{\max}],
    \label{eq:app_depth_cap}
\end{equation}
and obtain its reference point as
\begin{equation}
    \mathbf{q}^{w}_i
    =
    \mathbf{q}^{w}(\mathbf{u}_i,\widehat d_i).
    \label{eq:app_dir_point}
\end{equation}
When $f_i^{\mathrm{far}}=1$, the point at $100$ m is only a
\emph{capped geometric reference}; it is not interpreted as a physical
surface.

\subsubsection{Object-Anchor Geometry}
For an object anchor $a_i^o$, the landmark detector $LD(\cdot)$, implemented
with Grounded SAM~\cite{ren2024grounded}, localizes the queried object. Among
valid detections, we retain the highest-confidence mask $\mathbf{M}_i$ and its
bounding box. If the selected mask contains no pixel with positive depth, the
object anchor is discarded because neither its far-depth ratio nor its metric
geometry is defined.

Long-range masks require special treatment because a large fraction of their
pixels may lie beyond the reliable metric range. For a mask with at least one
positive-depth pixel, we define the far-depth ratio
\begin{equation}
\rho_i^{\mathrm{far}}
=
\frac{
\sum_{\mathbf{u}}
\mathbbm{1}
[
M_i(\mathbf{u})=1
\land
D_t(\mathbf{u})>D_{\max}
]
}{
\sum_{\mathbf{u}}
\mathbbm{1}
[
M_i(\mathbf{u})=1
\land
D_t(\mathbf{u})>0
]
}.
\label{eq:app_far_ratio}
\end{equation}
If
\begin{equation}
    \rho_i^{\mathrm{far}}
    >
    \tau_{\mathrm{far}},
    \qquad
    \tau_{\mathrm{far}}=0.5,
    \label{eq:app_object_degrade}
\end{equation}
more than half of the observed object support lies beyond the reliable range,
making its spatial extent unreliable. We therefore degrade the object anchor
into a directional anchor located at the mask centroid $\mathbf{u}^{c}_i$.
The resulting directional anchor is grounded using
Eqs.~\ref{eq:app_dir_depth}--\ref{eq:app_dir_point} and is not written into
the SOKB. This preserves useful directional information without introducing
inaccurate object geometry into persistent memory.

Otherwise, only mask pixels satisfying $0<D_t(\mathbf{u})\leq D_{\max}$ are
back-projected:
\begin{equation}
\begin{aligned}
\mathcal{Q}_i = \bigl\{
&\mathbf{q}^{w}(\mathbf{u}, D_t(\mathbf{u})) \mid
M_i(\mathbf{u}) = 1,\\
&0 < D_t(\mathbf{u}) \leq D_{\max}
\bigr\}.
\end{aligned}
\label{eq:app_object_points}
\end{equation}
If $\mathcal{Q}_i=\varnothing$ after reliable-depth filtering, the anchor is
discarded. Otherwise, the temporary 3D points are compressed into the 2.5D
geometry
\begin{equation}
    \mathbf{g}_i
    =
    (R_i,z_i^{\min},z_i^{\max}),
\end{equation}
where
\begin{align}
R_i
&=
\operatorname{ConvHull}
\left(
\Pi_{xy}(\mathcal{Q}_i)
\right),\\
z_i^{\min}
&=
\min_{\mathbf{q}\in\mathcal{Q}_i}q_z,
\qquad
z_i^{\max}
=
\max_{\mathbf{q}\in\mathcal{Q}_i}q_z.
\end{align}
Its reference center is
\begin{equation}
\mathbf{c}_i
=
\begin{bmatrix}
\operatorname{Centroid}(R_i)_x\\
\operatorname{Centroid}(R_i)_y\\
(z_i^{\min}+z_i^{\max})/2
\end{bmatrix}.
\label{eq:app_object_center}
\end{equation}
The temporary point cloud is discarded after constructing $\mathbf{g}_i$.

For a valid object anchor, we associate its detector confidence with the
reliability of its metric observation:
\begin{equation}
    q_i
    =
    q_i^{\mathrm{det}}
    \left(
    1-\rho_i^{\mathrm{far}}
    \right),
    \label{eq:app_anchor_conf}
\end{equation}
where $q_i^{\mathrm{det}}$ is the Grounded-SAM detection confidence. Thus,
partially far-range observations can still contribute to memory, but receive
lower reliability.

\subsubsection{Spatial Cue Computation}\label{app:spa_cue}
To avoid conflating the scalar observation reliability $q_i$ with geometric
points, we use the appendix-local notation $\mathbf{p}^{\mathrm{ref}}_i$ for
the world-coordinate reference point of an anchor:
\begin{equation}
\mathbf{p}^{\mathrm{ref}}_i
=
\begin{cases}
\mathbf{c}_i, & a_i\in\mathcal{A}_t^o,\\
\mathbf{q}^{w}_i, & a_i\in\mathcal{A}_t^d.
\end{cases}
\label{eq:app_anchor_reference}
\end{equation}
Let the UAV world position be $\mathbf{x}_t=(x_t,y_t,z_t)^\top$ with heading
$\psi_t$, and write
$\mathbf{p}^{\mathrm{ref}}_i=(p_{x,i},p_{y,i},p_{z,i})^\top$. Define
\begin{align}
\Delta x_{t,i}&=p_{x,i}-x_t,\\
\Delta y_{t,i}&=p_{y,i}-y_t,\\
\Delta z_{t,i}&=p_{z,i}-z_t.
\end{align}
Because the equations operate in the $z$-up world frame, $\Delta z_{t,i}>0$
means that the anchor lies above the UAV. We compute
\begin{align}
d^{xy}_{t,i}
&=
\sqrt{
\Delta x_{t,i}^{2}
+
\Delta y_{t,i}^{2}
},\\
d^{3D}_{t,i}
&=
\sqrt{
(d^{xy}_{t,i})^2
+
\Delta z_{t,i}^{2}
},
\end{align}
and the relative bearing
\begin{equation}
\phi_{t,i}
=
\operatorname{wrap}_{[-\pi,\pi)}
\left[
\operatorname{atan2}
(\Delta y_{t,i},\Delta x_{t,i})
-
\psi_t
\right].
\end{equation}
Hence,
\begin{equation}
\mathbf{r}_{t,i}
=
(
\phi_{t,i},
\Delta z_{t,i},
d^{xy}_{t,i},
d^{3D}_{t,i}
),
\end{equation}
which is shared by object and directional anchors.

\subsubsection{EAG Representation}
Each anchor node in $\mathcal{G}_t=(\mathcal{V}_t,\mathcal{E}_t)$ is represented together with the relative spatial cue encoded on its UAV-anchor edge. For an
object anchor, the corresponding edge is described, for example, as
\begin{quote}
\small
\texttt{Anchor 1 [Object: gray building]: The object center is 24.6 degrees to
your left, 3.2 m above the UAV, at a horizontal distance of 18.5 m and a 3D
distance of 18.8 m.}
\end{quote}
For a directional anchor, the corresponding edge is described as
\begin{quote}
\small
\texttt{Anchor 2 [Direction]: Free travel distance along this direction: 35.0 m.
The ray-cast endpoint is 12.4 degrees to your right, 6.8 m below the UAV, at a
horizontal distance of 34.3 m.}
\end{quote}
For a capped far-range directional anchor, we use
\begin{quote}
\small
\texttt{Anchor 3 [Direction]: Free travel distance along this direction exceeds 100
m. The geometric reference is capped at 100 m and lies 12.4 degrees to your
right, 8.5 m below the UAV.}
\end{quote}
The same anchor indices are overlaid on the RGB observation $I_t$ to obtain
$I'_t$, explicitly aligning the EAG with its visual evidence.

\subsection{Persistent Object Spatial Memory}
\label{app:memory}

\subsubsection{SOKB Representation}
The SOKB follows the type-instance hierarchy defined in the main paper:
\begin{equation}
\begin{aligned}
\mathcal{M}
&=
\{\mathcal{T}_m\}_{m=1}^{N_T},\\
\mathcal{T}_m
&=
(\mathcal{B}^{T}_m,\mathcal{O}_m),\\
\mathcal{O}_m
&=
\{o_j\}_{j=1}^{N_m},\\
o_j
&=
(\mathcal{B}^{A}_j,
\mathcal{B}^{V}_j,
\mathbf{g}_j,
\mathbf{c}_j,
q_j).
\end{aligned}
\label{eq:app_sokb}
\end{equation}
$\mathcal{B}^{T}_m$ contains alternative category expressions of object type
$\mathcal{T}_m$. $\mathcal{B}^{A}_j$ stores appearance descriptions and
$\mathcal{B}^{V}_j$ stores normalized DINOv2~\cite{oquab2024tmlr-dinov2} features
for instance $o_j$. $\mathbf{g}_j$ is its persistent 2.5D geometry,
$\mathbf{c}_j$ its derived center, and $q_j\in[0,1]$ its current reliability.
The center $\mathbf{c}_j$ is retained as a cached quantity for efficient
retrieval, but it is not independently fused: after every geometry update it
is recomputed deterministically from $\mathbf{g}_j$ using the same rule as
Eq.~\ref{eq:app_object_center}.

\subsubsection{Type-Level Matching}
We use GTE-small~\cite{li2023towards} to obtain text embeddings. Let
$E(\cdot)$ denote its $\ell_2$-normalized embedding. For an incoming category
label $\ell_i^T$, the similarity to type $\mathcal{T}_m$ is
\begin{equation}
S_T(i,m)
=
\max_{\ell\in\mathcal{B}^{T}_m}
E(\ell_i^T)^\top E(\ell).
\label{eq:app_type_sim}
\end{equation}
Candidate types are defined as
\begin{equation}
\mathcal{C}^{T}_i
=
\left\{
m
\mid
S_T(i,m)\geq\tau_T
\right\}.
\label{eq:app_type_candidates}
\end{equation}
If $\mathcal{C}^{T}_i=\varnothing$, a new type and its first instance are
initialized directly from $a_i^o$. Otherwise, we define the best compatible
type
\begin{equation}
m^*
=
\arg\max_{m\in\mathcal{C}^{T}_i}S_T(i,m),
\label{eq:app_best_type}
\end{equation}
which is used whenever the incoming observation is not associated with an
existing instance.

\subsubsection{Geometry Gate}
For incoming BEV footprint $R_i$ and stored footprint $R_j$, their minimum
horizontal separation is
\begin{equation}
d^{\mathrm{BEV}}_{ij}
=
\begin{cases}
0,
&
R_i\cap R_j\neq\varnothing,\\[1mm]
\displaystyle
\min_{\mathbf{x}\in R_i,\mathbf{y}\in R_j}
\|\mathbf{x}-\mathbf{y}\|_2,
&
\text{otherwise}.
\end{cases}
\label{eq:app_bev_distance}
\end{equation}
Instances satisfying
\begin{equation}
    d^{\mathrm{BEV}}_{ij}>D_G
\end{equation}
are removed before fine-grained matching. We use $D_G=20$ m. Equivalently,
the gated candidate set is
\begin{equation}
\mathcal{J}_i
=
\left\{
j
\,\middle|\,
m(j)\in\mathcal{C}^{T}_i,
\ d^{\mathrm{BEV}}_{ij}\leq D_G
\right\},
\label{eq:app_instance_candidates}
\end{equation}
where $m(j)$ denotes the parent type of $o_j$.

\subsubsection{Fine-Grained Object Association}
For candidates passing the geometry gate, geometric similarity combines
observation-to-memory footprint coverage and horizontal distance:
\begin{equation}
S_{\mathrm{cov}}(i,j)
=
\frac{
\operatorname{Area}(R_i\cap R_j)
}{
\operatorname{Area}(R_i)+\epsilon
},
\qquad \epsilon=10^{-6},
\label{eq:app_coverage}
\end{equation}
\begin{equation}
S_{\mathrm{dist}}(i,j)
=
\exp
\left(
-\frac{d^{\mathrm{BEV}}_{ij}}{\sigma_G}
\right),
\end{equation}
and
\begin{equation}
S_G(i,j)
=
\lambda_GS_{\mathrm{cov}}(i,j)
+
(1-\lambda_G)S_{\mathrm{dist}}(i,j).
\label{eq:app_geo_sim}
\end{equation}
The coverage term in Eq.~\ref{eq:app_coverage} is intentionally asymmetric:
it measures how much of the newly observed footprint $R_i$ is explained by
the persistent footprint $R_j$. This is preferable to a symmetric IoU here
because a single aerial observation may cover only a partial region of a
large object whose persistent geometry has already accumulated multiple
views.

When a non-empty appearance description $\ell_i^A$ and at least one stored
appearance description are available, appearance-level semantic similarity is
\begin{equation}
S_A(i,j)
=
\max_{\ell\in\mathcal{B}^{A}_j}
E(\ell_i^A)^\top E(\ell).
\end{equation}
The semantic score is then
\begin{equation}
S_S(i,j)
=
\lambda_S S_T(i,m(j))
+
(1-\lambda_S)S_A(i,j).
\label{eq:app_sem_sim}
\end{equation}
If the incoming object has no intrinsic appearance attribute, or if
$\mathcal{B}^{A}_j$ is empty, the unavailable appearance term is omitted and
we set $S_S(i,j)=S_T(i,m(j))$. This convention avoids embedding an empty
string while preserving the category score as the complete semantic signal.

For visual matching, let $\mathbf{v}_i$ denote the normalized DINOv2 embedding
of the current object crop. We compute
\begin{equation}
S_V(i,j)
=
\max_{\mathbf{v}\in\mathcal{B}^{V}_j}
\mathbf{v}_i^\top\mathbf{v}.
\label{eq:app_vis_sim}
\end{equation}
The final association score is
\begin{equation}
\begin{split}
S_M(i,j)
={}&
w_GS_G(i,j)
+
w_SS_S(i,j)\\
&+
w_VS_V(i,j),
\end{split}
\label{eq:app_match_score}
\end{equation}
where $w_G+w_S+w_V=1$. The best gated candidate is
\begin{equation}
    j^*
    =
    \arg\max_{j\in\mathcal{J}_i} S_M(i,j).
    \label{eq:app_best_instance}
\end{equation}
If $S_M(i,j^*)\geq\tau_M$, the incoming anchor is fused with $o_{j^*}$;
otherwise, a new instance is created under the best compatible type
$\mathcal{T}_{m^*}$ from Eq.~\ref{eq:app_best_type}. If
$\mathcal{J}_i=\varnothing$, the fine-grained match is skipped and a new
instance is also created under $\mathcal{T}_{m^*}$.

\subsubsection{Geometry and Attribute Fusion}
When $a_i^o$ is associated with $o_j$, the persistent geometry is updated as
\begin{align}
R_j
&\leftarrow
\operatorname{ConvHull}(R_j\cup R_i),\\
z_j^{\min}
&\leftarrow
\min(z_j^{\min},z_i^{\min}),\\
z_j^{\max}
&\leftarrow
\max(z_j^{\max},z_i^{\max}).
\end{align}
The center $\mathbf{c}_j$ is then recomputed using
Eq.~\ref{eq:app_object_center}. The instance confidence jointly incorporates
the reliability of the new observation and its association consistency:
\begin{equation}
q_j
\leftarrow
(1-\alpha_q)q_j
+
\alpha_q
\left[
q_i S_M(i,j)
\right].
\label{eq:app_conf_update}
\end{equation}
A newly created instance is initialized with $q_j=q_i$. Thus, a
high-confidence observation that is strongly consistent with the stored
object reinforces the persistent instance, whereas a marginal association
produces only a small update.

The category bank $\mathcal{B}^T_{m(j)}$, appearance bank
$\mathcal{B}^A_j$, and visual bank $\mathcal{B}^V_j$ are all bounded. Each
stored bank entry retains the reliability of the observation that produced it
solely for bank replacement; for an entry $b$, $q(b)$ denotes this inherited
reliability. Whenever a new entry causes a bank to exceed its capacity, we
compute
\begin{equation}
U(b)
=
\lambda_Bq(b)
+
(1-\lambda_B)
\min_{b'\neq b}
\left[
1-\operatorname{sim}(b,b')
\right]
\label{eq:app_bank_utility}
\end{equation}
for every entry in the temporarily expanded bank and remove the entry with
the lowest utility. We use cosine similarity of GTE-small embeddings for the
category/appearance banks and cosine similarity of normalized DINOv2 features
for the visual bank. This update requires no separate redundancy threshold:
near-identical repeated observations naturally receive low diversity utility,
while reliable evidence from distinct aerial viewpoints is retained. For a
successful association, the incoming category expression, non-empty
appearance description, and visual feature are respectively considered for
$\mathcal{B}^T_{m(j)}$, $\mathcal{B}^A_j$, and $\mathcal{B}^V_j$ using the
same bounded-bank rule.

\subsubsection{Landmark Prior Retrieval}
For landmark $\ell$ in the active subtask, we represent its category and
appearance description as $\ell^T$ and $\ell^A$. Define the category
compatibility
\begin{equation}
S_R^T(\ell,o_j)
=
\max_{l\in\mathcal{B}^{T}_{m(j)}}
E(\ell^T)^\top E(l).
\end{equation}
When $\ell^A$ is non-empty and $\mathcal{B}^A_j$ contains at least one
appearance description, define
\begin{equation}
S_R^A(\ell,o_j)
=
\max_{l\in\mathcal{B}^{A}_{j}}
E(\ell^A)^\top E(l),
\end{equation}
and use
\begin{equation}
S_R(\ell,o_j)
=
\lambda_R S_R^T(\ell,o_j)
+
(1-\lambda_R)S_R^A(\ell,o_j).
\label{eq:app_retrieval_sem}
\end{equation}
If no appearance constraint is present in the instruction, or if the instance
has no stored appearance description, the unavailable appearance term is
omitted and $S_R(\ell,o_j)=S_R^T(\ell,o_j)$.

We use the stored instance reliability to softly calibrate candidate ranking:
\begin{equation}
\widetilde S_R(\ell,o_j)
=
S_R(\ell,o_j)
\left[
\beta_q
+
(1-\beta_q)q_j
\right].
\label{eq:app_conf_retrieval}
\end{equation}
Because $\beta_q$ is close to one, semantic compatibility remains the
dominant signal; confidence acts only as supporting evidence between
otherwise comparable instances. Candidates satisfy
\begin{equation}
S_R(\ell,o_j)\geq\tau_R,
\qquad
d^{xy}_{t,j}\leq D_R.
\end{equation}
The top-$K_R$ instances according to $\widetilde S_R$ form
$\mathcal{C}_{\ell}$. For each candidate, its UAV-relative spatial cue
$\mathbf{r}_{t,j}$ is recomputed from its persistent center. We define
\begin{equation}
\mathcal{R}_{t,\ell}
=
\left\{
(o_j,\mathbf{r}_{t,j},q_j)
\mid
o_j\in\mathcal{C}_{\ell}
\right\}.
\end{equation}
The MLLM selector jointly reasons over the active subtask and the candidate
sets $\mathcal{R}_{t,\ell}$ for all $\ell\in\mathcal{L}_n$, and may select one
instance or return \textit{None} for each landmark. Once selected, the
identity of a landmark remains fixed within the active subtask, while its
relative cue is recomputed as the UAV moves. The selected instances and
current cues constitute the global prior $\mathcal{P}_t$.

\subsubsection{Memory Lifecycle During Evaluation}
\label{app:memory_protocol}

AirAnchor adopts a scene-level online-persistent memory protocol.
A separate SOKB is maintained for each environment scene, implemented
as a dictionary indexed by the scene identifier. The SOKB of a scene
is initialized as empty when the first evaluation episode from that
scene is encountered and is subsequently updated using object anchors
acquired during navigation.

Unless otherwise specified, all experiments follow the canonical
episode order provided by the corresponding AerialVLN-S split.
Episodes are not regrouped by scene. A separate SOKB is maintained for
each scene; when evaluation later returns to a previously encountered
scene, its accumulated SOKB is restored and continues to be updated.
Accordingly, each episode can access only object knowledge accumulated
from earlier evaluated episodes of the same scene under the current
episode ordering. No information from future episodes is available. All online memory updates use only object anchors obtained from RGB-D
observations and UAV poses actually collected by the agent.

\subsection{Spatially-Informed Navigation Agent}
\label{app:agent}

\subsubsection{Subtask Management}
At the beginning of each episode, $\mathrm{MLLM}_{\mathrm{dec}}$ decomposes
the complete instruction into
\begin{equation}
    \{(s_n,\mathcal{L}_n,\mathcal{U}_n)\}_{n=1}^{N}.
\end{equation}
Each $s_n$ preserves the original execution order and relational language,
while $\mathcal{L}_n$ contains the landmarks used to query the SOKB. Only one
subtask is active at a time, and landmark retrieval is repeated when the agent
advances to a new subtask. Each subtask is further decomposed into a non-empty
ordered list of subgoals $\mathcal{U}_n$, used only to compute the
automatic-backtracking threshold
\begin{equation}
    B_n=\lambda_{\mathrm{step}}|\mathcal{U}_n|,
    \qquad \lambda_{\mathrm{step}}=3.
\end{equation}

The controller counts agent-loop iterations for the active subtask since its
start or most recent backtracking. If this count exceeds $B_n$ while the
subtask remains incomplete, it automatically triggers Path Backtracking,
subject to a shared limit of two backtracking executions per subtask for both
agent-selected and automatic recovery. The local iteration counter is reset
after backtracking; the backtracking count is reset only for a new subtask.
Once the backtracking limit is reached, further backtracking is disabled.

\subsubsection{Skill-Level Planning}\label{app:skill}
A parameterized skill is denoted by
\begin{equation}
    \sigma_t
    =
    k_t(\boldsymbol{\eta}_t).
\end{equation}
The navigation MLLM predicts $(\sigma_t,\xi_t)$ from the visually prompted
observation, navigation context, EAG, and global landmark priors. The skill
parameters are produced in the same navigation call; no additional MLLM call
is used to separately predict Pixel-Navigation distance or Altitude-Adjustment
magnitude. We implement four skills tailored to aerial navigation.

\noindent\textbf{Pixel Navigation.}
Pixel Navigation follows the image-space waypoint formulation of
SPF~\cite{hu2025see}. Its parameter is
\begin{equation}
\boldsymbol{\eta}^{\mathrm{pix}}_t
=
(u_t,v_t,d_t),
\end{equation}
where $(u_t,v_t)$ is an image-space waypoint and $d_t$ is its metric travel
depth. AirAnchor explicitly provides metric anchor cues, allowing the MLLM to
predict $d_t$ in meters. Let
\begin{equation}
\widehat{\mathbf{r}}^{c}_t
=
\frac{
\mathbf{K}^{-1}(u_t,v_t,1)^\top
}{
\left\|
\mathbf{K}^{-1}(u_t,v_t,1)^\top
\right\|_2
}
\end{equation}
be the normalized camera ray. Its world-frame direction is
\begin{equation}
\widehat{\mathbf{r}}^{w}_t
=
\mathbf{R}^{wc}_t
\widehat{\mathbf{r}}^{c}_t.
\end{equation}
The target position is
\begin{equation}
\mathbf{x}^{*}_t
=
\mathbf{x}_t
+
d_t
\widehat{\mathbf{r}}^{w}_t.
\label{eq:app_pixel_goal}
\end{equation}

\noindent\textbf{Altitude Adjustment.}
Altitude Adjustment receives a signed metric vertical displacement
\begin{equation}
\boldsymbol{\eta}^{\mathrm{alt}}_t
=
\Delta h_t,
\end{equation}
and produces
\begin{equation}
\mathbf{x}^{*}_t
=
(x_t,y_t,z_t+\Delta h_t)^\top.
\end{equation}
Because all spatial computation uses the $z$-up world frame,
$\Delta h_t>0$ denotes ascent and $\Delta h_t<0$ denotes descent.

\noindent\textbf{View Rotation.}
View Rotation performs an in-place panoramic observation to support
fine-grained heading estimation beyond the current field of view.
The UAV position remains fixed, and the heading at the beginning of
the scan is retained as the reference heading. Eight RGB observations
are acquired at uniformly distributed relative yaw offsets:
\begin{equation}
\begin{aligned}
\{\delta\psi_m\}_{m=0}^{7}
=
\{&
0^\circ,45^\circ,90^\circ,135^\circ,180^\circ,\\
&-135^\circ,-90^\circ,-45^\circ
\}.
\end{aligned}
\label{eq:app_pan_yaws}
\end{equation}
The resulting panoramic observation is
\begin{equation}
\mathcal{I}^{\mathrm{pan}}_t
=
\left\{
\left(I^{(m)}_t,\delta\psi_m\right)
\right\}_{m=0}^{7}.
\end{equation}

The discrete views serve as panoramic visual references rather than
candidate actions. Given the eight observations and their signed yaw
offsets, the MLLM first identifies the coarse direction that best
supports the active subtask and then performs fine-grained angular
reasoning between the observed directions. It predicts a target
relative yaw
\begin{equation}
\Delta\psi_t^{\mathrm{pred}}
\in[-180^\circ,180^\circ],
\end{equation}
which is not restricted to the eight sampled yaw offsets.

Because the environment provides a $15^\circ$ rotation primitive, the
predicted angle is converted to the closest executable rotation:
\begin{equation}
\Delta\psi_t^{\mathrm{exec}}
=
15^\circ
\operatorname{round}
\left(
\frac{\Delta\psi_t^{\mathrm{pred}}}{15^\circ}
\right).
\label{eq:app_view_quantization}
\end{equation}
The UAV then rotates from the scan-reference heading by
$\Delta\psi_t^{\mathrm{exec}}$. View Rotation changes only the UAV
heading and does not invoke the local A$^\ast$ planner.

\noindent\textbf{Path Backtracking.}
Path Backtracking operates exclusively on the execution history of the current
subtask. Directional anchors are not retained in history because they
represent transient geometry useful only for the decision at which they were
grounded. Let
\begin{equation}
    \mathcal{G}^{o}_t
    \subseteq
    \mathcal{G}_t
\end{equation}
denote the EAG obtained by retaining only the UAV node, object-anchor nodes,
and corresponding UAV--object edges. The current-subtask history is organized
as a path chain
\begin{equation}
    \mathcal{H}_n
    =
    (\mathcal{V}^{H}_n,\mathcal{E}^{H}_n).
\end{equation}
Each history node is
\begin{equation}
h_j
=
\left(
\mathbf{p}_j,
\mathcal{C}_j,
\chi_j,
\mathcal{G}^{o}_j
\right),
\label{eq:app_history_node}
\end{equation}
where $\chi_j$ is a concise scene caption generated in the same MLLM call as
the corresponding skill-level decision. The edge between two consecutive
history nodes is
\begin{equation}
e_j
=
\left(
\sigma_j,
\xi_j,
\Delta\mathbf{p}_j
\right),
\label{eq:app_history_edge}
\end{equation}
where the realized displacement is
\begin{equation}
\Delta\mathbf{p}_j
=
\left(
\mathbf{x}_{j+k}-\mathbf{x}_j,\,
\operatorname{wrap}(\psi_{j+k}-\psi_j)
\right).
\end{equation}
Thus, an edge records both the skill-level decision and the motion that was
actually executed. Path Backtracking can be selected by the agent or triggered
automatically when the subtask iteration count exceeds its threshold, with
at most two executions per subtask in total. In either case, the MLLM receives the
current subtask and serialized history chain $\mathcal{H}_n$ and selects
\begin{equation}
j_b
=
\mathrm{MLLM}_{\mathrm{back}}
(s_n,\mathcal{H}_n).
\end{equation}
The executor then retraces the previously executed path-chain transitions from
the current state back to node $h_{j_b}$. Since these transitions have already
been traversed in the same scene, backtracking reuses the stored
transition sequence and does not invoke the local A$^\ast$ planner.

\subsubsection{Local \texorpdfstring{A$^\ast$}{A*} Execution}
\label{app:local_astar}
A$^\ast$~\cite{hart1968formal} is invoked only for \textit{Pixel Navigation}
and \textit{Altitude Adjustment}. It does not access a preconstructed scene
map, simulator navigation graph, or reference trajectory. Instead, a
temporary local free-space representation is constructed solely from the
depth observation associated with the selected skill.

Consistent with Appendix~\ref{app:depth_grounding}, all depth measurements are
capped at
\begin{equation}
    \widehat D(\mathbf{u})
    =
    \min(D(\mathbf{u}),D_{\max}).
\end{equation}
For a viewing ray $\widehat{\mathbf r}(\mathbf{u})$, its locally observed free
segment is
\begin{equation}
\begin{aligned}
\mathcal{F}(\mathbf{u})
=
\bigl\{
&\mathbf{x}_t
+s\,\widehat{\mathbf r}(\mathbf{u})
\;\big|\; \\
&0<s<
\max\bigl(
0,\widehat D(\mathbf{u})-\delta_{\mathrm{safe}}
\bigr)
\bigr\}.
\end{aligned}
\end{equation}
When the original depth exceeds $D_{\max}$, the ray is therefore regarded as
free only up to the reliable sensing boundary rather than being assumed
obstacle-free indefinitely.

For Pixel Navigation, the local representation is constructed from the
current egocentric depth $D_t$, and the commanded metric depth is constrained
by
\begin{equation}
d_t^{\mathrm{exec}}
=
\max\left(
0,
\min\left(
d_t,\,
\widehat D_t(u_t,v_t)-\delta_{\mathrm{safe}}
\right)
\right).
\label{eq:app_exec_depth}
\end{equation}
If $d_t^{\mathrm{exec}}=0$, the translational request is treated as locally
infeasible. Otherwise, A$^\ast$ searches the temporary local action lattice
toward the resulting waypoint. For Altitude Adjustment, we obtain an
upward-facing depth observation when ascending and a downward-facing depth
observation when descending, and construct the local free-space
representation in the same manner. View Rotation is executed directly
through yaw rotations, whereas Path Backtracking retraces the stored history
chain and does not invoke A$^\ast$.

If no feasible local path is found, the executor does not substitute an oracle
waypoint. The failed transition is recorded in the navigation context and
control returns to the high-level planner for a new closed-loop decision.

\subsubsection{Progress Reflection}
Suppose skill $\sigma_t$ executes $k$ primitive actions and moves the UAV from
$\mathbf{p}_t$ to $\mathbf{p}_{t+k}$. The world-coordinate reference point
$\mathbf{p}^{\mathrm{ref}}_i$ of every anchor in the decision-time EAG
$\mathcal{G}_t$ is retained until reflection. Its relative cue is recomputed
at the new pose:
\begin{equation}
\widetilde{\mathbf{r}}_{t+k,i}
=
\operatorname{Rel}
\left(
\mathbf{p}_{t+k},
\mathbf{p}^{\mathrm{ref}}_i
\right),
\end{equation}
where $\operatorname{Rel}(\cdot)$ denotes the same relative-geometry
computation used in Appendix~\ref{app:anchor}. The recentered graph is
\begin{equation}
\begin{aligned}
\widetilde{\mathcal{G}}_{t+k}
&=
(\widetilde{\mathcal{V}}_{t+k},
\widetilde{\mathcal{E}}_{t+k}),\\
\widetilde{\mathcal{V}}_{t+k}
&=
\{v^u_{t+k}\}
\cup
\{v_i\mid a_i\in\mathcal{A}_t\},\\
\widetilde{\mathcal{E}}_{t+k}
&=
\left\{
(v^u_{t+k},v_i,
\widetilde{\mathbf{r}}_{t+k,i})
\,\middle|\,
a_i\in\mathcal{A}_t
\right\}.
\end{aligned}
\label{eq:app_recentered_graph}
\end{equation}
Thus, $\mathcal{G}_t$ and $\widetilde{\mathcal{G}}_{t+k}$ contain the same
anchor identities but express their relations to the UAV before and after
skill execution. Their comparison explicitly exposes changes such as
approaching, passing, crossing, or flying above a landmark.

Following the notation in the main paper,
\begin{align}
\mathcal{C}^{\mathrm{sem}}_t
&=
(\mathcal{C}_t,I'_t,I'_{t+k},\sigma_t,\xi_t),\\
\mathcal{C}^{\mathrm{spa}}_t
&=
(\mathcal{G}_t,\widetilde{\mathcal{G}}_{t+k}).
\end{align}
The reflection MLLM predicts
\begin{equation}
(\gamma_{t+k},b_{t+k},\pi_{t+k})
=
\mathrm{MLLM}_{\mathrm{ref}}
(\mathcal{C}^{\mathrm{sem}}_t,
\mathcal{C}^{\mathrm{spa}}_t).
\end{equation}
If $b_{t+k}$ indicates completion, the agent advances to the next subtask and
retrieves its corresponding landmark priors. To avoid carrying the completed
subtask state into the next one, the progress variable is reset to
\textsc{NotStarted} and the plan is initialized from the newly active subtask
text, without using its auxiliary subgoals. Completion of the final subtask
triggers \textsc{Stop}; otherwise, the episode terminates when the global
budget of 20 agent-loop iterations is exhausted.

\subsection{Hyperparameter Settings}
\label{app:hyperparameters}
Table~\ref{tab:app_hyperparameters} summarizes the hyperparameters used in
AirAnchor.

\begin{table*}[t]
\centering
\small
\setlength{\tabcolsep}{5.2pt}
\renewcommand{\arraystretch}{1.08}
\begin{tabular}{@{}llcp{7.0cm}@{}}
\toprule
\textbf{Module}
& \textbf{Parameter}
& \textbf{Value}
& \textbf{Description}\\
\midrule

Observation
& RGB resolution
& $512\times512$
& Resolution of RGB observations.\\
& Depth resolution
& $512\times512$
& Resolution of depth observations.\\

\midrule
Anchor
& Queried anchors $K$
& $3$
& Number of spatial anchors per decision step.\\
& Directional depth window
& $5\times5$
& Neighborhood used for directional depth aggregation.\\
& Reliable depth $D_{\max}$
& $100$ m
& Maximum depth.\\
& Far-object threshold $\tau_{\mathrm{far}}$
& $0.50$
& Object-to-directional fallback threshold in Eq.~\ref{eq:app_object_degrade}.\\
& Grounded-SAM box / text threshold
& $0.30/0.25$
& Open-vocabulary object-grounding thresholds.\\
\midrule
Memory
& Bank capacities
& $3/3/3$
& Type-label / appearance-label / visual-feature capacities.\\
& Type threshold $\tau_T$
& $0.55$
& Minimum similarity for compatible object types.\\
& Geometry gate $D_G$
& $20$ m
& Maximum BEV separation for instance association.\\
& $\lambda_G,\sigma_G$
& $0.5,\ 10$ m
& Observation-coverage / distance geometric-similarity.\\
& $\lambda_S$
& $0.5$
& Category--appearance semantic trade-off.\\
& $(w_G,w_S,w_V)$
& $(0.40,0.35,0.25)$
& Geometry / semantics / vision association weights.\\
& Match threshold $\tau_M$
& $0.65$
& Minimum score for merging an incoming anchor.\\
& Confidence rate $\alpha_q$
& $0.20$
& Instance-reliability update rate.\\
& Bank utility weight $\lambda_B$
& $0.50$
& Reliability-diversity trade-off for bank replacement.\\
& Retrieval semantic weight $\lambda_R$
& $0.50$
& Category and appearance trade-off.\\
& Confidence floor $\beta_q$
& $0.80$
& Confidence calibration strength for retrieval ranking.\\
& Retrieval threshold $\tau_R$
& $0.50$
& Minimum semantic compatibility for landmark retrieval.\\
& Retrieval radius $D_R$
& $80$ m
& Maximum horizontal distance of retrieval candidates.\\
& Retrieval top-$K_R$
& $3$
& Maximum candidates per landmark.\\
\midrule
Agent
& $\lambda_{\mathrm{step}}$
& $3$
& Multiplier for the automatic-backtracking threshold\\
& Backtracking limit
& $2$
& Maximum number of backtracking attempts per subtask.\\
& Episode agent-loop budget
& $20$
& Maximum number of agent-loop iterations per episode.\\
& Pixel travel range
& $[5,50]$ m
& Metric travel-distance range for Pixel Navigation.\\
& Safety margin $\delta_{\mathrm{safe}}$
& $2$ m
& Clearance from observed depth surfaces.\\
& Maximum altitude change
& $30$ m
& Maximum magnitude of one Altitude Adjustment.\\

\bottomrule
\end{tabular}
\caption{Implementation settings of AirAnchor.}
\label{tab:app_hyperparameters}
\end{table*}

\subsection{Pseudocode}
\label{app:algorithm}
Algorithm~\ref{alg:airanchor} summarizes the complete navigation pipeline,
and Algorithm~\ref{alg:sokb} details incremental SOKB maintenance.
Here, $\ell$ counts all agent-loop iterations, $c$ counts iterations since the
current subtask began or last backtracked, and $r$ counts backtracking
executions within that subtask. The skill set $\mathcal{S}_t$ constrains the
available choices without changing the subtask input. If retrieval returns
no candidates, the global prior is empty. Skill execution follows
Appendix~\ref{app:skill}, including panoramic reasoning with global priors,
history-based recovery-node selection, and the local planner where applicable.
The automatic trigger is checked before the next skill execution after
$c>B_n$; completion and the global budget take precedence over recovery.

\begin{algorithm*}[t]
\small
\caption{AirAnchor Navigation}
\label{alg:airanchor}
\begin{algorithmic}[1]
\Require Instruction $\mathcal{I}$, scene identifier $\mathrm{sid}$, scene-indexed memory store $\mathfrak{M}$
\Ensure Navigation trajectory $\mathcal{T}$ and updated memory store $\mathfrak{M}$
\State $\mathcal{M}\gets\mathfrak{M}[\mathrm{sid}]$ if available; otherwise $\mathcal{M}\gets\varnothing$
\State $\{(s_n,\mathcal{L}_n,\mathcal{U}_n)\}_{n=1}^{N}\gets\mathrm{MLLM}_{\mathrm{dec}}(\mathcal{I})$
\State $t\gets0$; $\ell\gets0$; initialize $\mathcal{T}$ with the initial UAV pose
\For{$n=1,\ldots,N$}
    \If{$\ell=20$} \State \textbf{break} \EndIf
    \State $B_n\gets3|\mathcal{U}_n|$; $c\gets0$; $r\gets0$; $\mathcal{H}_n\gets\varnothing$
    \State $\gamma_t\gets\textsc{NotStarted}$; initialize $\pi_t$ from $s_n$; $\mathcal{C}_t\gets(s_n,\gamma_t,\pi_t)$
    \State Read current pose; retrieve candidates for $\mathcal{L}_n$ and select identities with $\mathrm{MLLM}_{\mathrm{ret}}$ if non-empty
    \While{$\ell<20$}
        \State Observe $(I_t,D_t,\mathbf{p}_t)$; construct $\mathcal{P}_t$ by recentering the selected landmark priors at $\mathbf{p}_t$
        \State $\mathcal{A}_t\gets\mathrm{MLLM}_{\mathrm{query}}(I_t,\mathcal{C}_t)$
        \State $(\mathcal{G}_t,I'_t,\mathcal{A}^{o,\mathrm{valid}}_t)\gets\Call{GroundAnchors}{\mathcal{A}_t,D_t,\mathbf{p}_t,D_{\max}}$
        \ForAll{$a_i^o\in\mathcal{A}^{o,\mathrm{valid}}_t$}
            \State $\mathcal{M}\gets\Call{UpdateSOKB}{\mathcal{M},a_i^o}$
        \EndFor
        \State $\mathcal{S}_t\gets$ available skills; exclude Path Backtracking if $r=2$ or no previous recovery node exists
        \State $(\sigma_t,\xi_t,\chi_t)\gets\mathrm{MLLM}_{\mathrm{nav}}(I'_t,\mathcal{C}_t,\mathcal{G}_t,\mathcal{P}_t;\mathcal{S}_t)$
        \If{$c>B_n$ and Path Backtracking $\in\mathcal{S}_t$}
            \State $\sigma_t\gets\textsc{PathBacktracking}(\varnothing)$; $\xi_t\gets$ automatic subtask-threshold trigger
        \EndIf
        \State Record decision node $h_t=(\mathbf{p}_t,\mathcal{C}_t,\chi_t,\mathcal{G}_t^o)$ in $\mathcal{H}_n$
        \State Execute $\sigma_t$ as specified in Appendix~\ref{app:skill}; obtain $(I_{t+k},\mathbf{p}_{t+k})$
        \State Append $(\sigma_t,\xi_t,\Delta\mathbf{p}_t)$ to $\mathcal{H}_n$ and the executed path to $\mathcal{T}$
        \State $\widetilde{\mathcal{G}}_{t+k}\gets\Call{Recenter}{\mathcal{G}_t,\mathbf{p}_{t+k}}$
        \State $(\gamma_{t+k},b_{t+k},\pi_{t+k})\gets\mathrm{MLLM}_{\mathrm{ref}}(\mathcal{C}^{\mathrm{sem}}_t,\mathcal{C}^{\mathrm{spa}}_t)$
        \State $\ell\gets\ell+1$; $c\gets c+1$
        \If{$\sigma_t$ is Path Backtracking}
            \State $r\gets r+1$; $c\gets0$
        \EndIf
        \State $t\gets t+k$; $\mathcal{C}_t\gets(s_n,\gamma_t,\pi_t)$
        \If{$b_t=\textsc{Completed}$}
            \State \textbf{break} \Comment{Advance to the next subtask, if any}
        \EndIf
    \EndWhile
\EndFor
\State Execute \textsc{Stop}; $\mathfrak{M}[\mathrm{sid}]\gets\mathcal{M}$
\State \Return $\mathcal{T}$ and $\mathfrak{M}$
\end{algorithmic}
\end{algorithm*}

\begin{algorithm}[t]
\small
\caption{Incremental SOKB Update}
\label{alg:sokb}
\begin{algorithmic}[1]
\Require Memory $\mathcal{M}$, valid object anchor $a_i^o$
\Ensure Updated $\mathcal{M}$
\State Compute $S_T(i,m)$ for all types
\State $\mathcal{C}^{T}_i\gets\{m:S_T(i,m)\geq\tau_T\}$
\If{$\mathcal{C}^{T}_i=\varnothing$}
    \State Create a new type and initialize its first instance from $a_i^o$
    \State \Return $\mathcal{M}$
\EndIf
\State $m^*\gets\arg\max_{m\in\mathcal{C}^{T}_i}S_T(i,m)$
\State $\mathcal{J}_i\gets\varnothing$
\ForAll{$o_j$ with $m(j)\in\mathcal{C}^{T}_i$}
    \If{$d^{\mathrm{BEV}}_{ij}\leq D_G$}
        \State $\mathcal{J}_i\gets\mathcal{J}_i\cup\{j\}$
    \EndIf
\EndFor
\If{$\mathcal{J}_i=\varnothing$}
    \State Create a new instance under $\mathcal{T}_{m^*}$
    \State \Return $\mathcal{M}$
\EndIf
\ForAll{$j\in\mathcal{J}_i$}
    \State Compute $S_G(i,j)$, $S_S(i,j)$, and $S_V(i,j)$
    \State Compute $S_M(i,j)$ using Eq.~\ref{eq:app_match_score}
\EndFor
\State $j^*\gets\arg\max_{j\in\mathcal{J}_i} S_M(i,j)$
\If{$S_M(i,j^*)\geq\tau_M$}
    \State Merge the persistent 2.5D geometry and recompute $\mathbf{c}_{j^*}$
    \State Update $\mathcal{B}^T_{m(j^*)}$, $\mathcal{B}^A_{j^*}$, and $\mathcal{B}^V_{j^*}$ using Eq.~\ref{eq:app_bank_utility}
    \State Update $q_{j^*}$ using Eq.~\ref{eq:app_conf_update}
\Else
    \State Create a new instance under $\mathcal{T}_{m^*}$
\EndIf
\State \Return $\mathcal{M}$
\end{algorithmic}
\end{algorithm}

\section{Additional Experimental Results}
\label{app:additional_exp}

\subsection{Additional Quantitative Analysis}
\label{app:quantitative}
All experiments in this subsection are conducted on the AerialVLN-S
Validation Seen split under the same evaluation protocol as the main
experiments.

\subsubsection{Impact of the Number of Spatial Anchors}
We analyze how the number of spatial anchors affects navigation performance.
A small $K$ provides insufficient spatial references, whereas an excessively
large $K$ may introduce redundant or less relevant information: the anchor
query gradually covers less decision-relevant objects, detection and spatial
grounding introduce additional noisy cues, and the enlarged EAG increases the
structured context presented to the MLLM, potentially diluting its attention
to critical spatial information. As shown in
Table~\ref{tab:anchor_sensitivity}, using only one anchor reduces SR from 9.6
to 8.4 and increases NE by 9.7\,m, indicating that a single reference is often
insufficient to capture complementary object-level and directional geometry.
Increasing $K$ to six also degrades performance, despite grounding more scene
elements. We therefore set $K=3$, which provides a compact set of
complementary spatial references while avoiding excessive redundancy and
noise in the EAG.

\begin{table}[t]
\centering
\small
\setlength{\tabcolsep}{7.5pt}
\begin{tabular}{@{}cccc@{}}
\toprule
$K$
& SR$\uparrow$
& SDTW$\uparrow$
& NE$\downarrow$\\
\midrule
1 & 8.4 & 3.3 & 90.2\\
\textbf{3}
& \textbf{9.6}
& \textbf{4.0}
& \textbf{80.5}\\
6 & 9.2 & 3.6 & 85.4\\
\bottomrule
\end{tabular}
\caption{Impact of the number of spatial anchors.}
\label{tab:anchor_sensitivity}
\end{table}

\subsubsection{Memory Construction Mechanism}
\label{app:memory_construction}

We further compare practical strategies for constructing the
scene-level global object memory. Besides the default online-persistent
SOKB, we construct a pre-rendered initialization inspired by offline
scene-memory construction~\cite{ning2026lookasidevln}. For each scene,
we uniformly sample 32 collision-free aerial poses from valid simulator
free space using a fixed sampling procedure. At each pose, four RGB-D
observations are rendered at yaw offsets of
$0^\circ$, $90^\circ$, $180^\circ$, and $270^\circ$, yielding
128 observations per scene. Since no navigation context is available
during pre-rendering, the MLLM receives only the RGB observation and
proposes up to three persistent, spatially stable, and visually
distinctive landmark candidates. Their geometry is obtained from the
corresponding depth observations, and the resulting object anchors are
consolidated using the same grounding, reliability estimation, and
SOKB association mechanisms as online AirAnchor observations.

We compare three settings:
\textit{Pre-rendered Init. (Frozen)} uses the pre-rendered SOKB
throughout evaluation without test-time updates;
\textit{Online Persistent} is the default AirAnchor setting, where
the SOKB of each scene starts empty and incrementally accumulates
query-driven object anchors across episodes from that scene; and
\textit{Pre-rendered Init. + Online Update} initializes the SOKB from
the pre-rendered observations and subsequently enables the same online
updates as AirAnchor.

\begin{table}[t]
\centering
\small
\setlength{\tabcolsep}{4.8pt}
\begin{tabular}{@{}lccc@{}}
\toprule
\textbf{Memory Construction}
& \textbf{SR}$\uparrow$
& \textbf{SDTW}$\uparrow$
& \textbf{NE}$\downarrow$\\
\midrule
Pre-rendered Init. (Frozen)
& 8.7 & 3.4 & 88.7\\
Online Persistent (Ours)
& 9.6 & 4.0 & 80.5\\
Pre-rendered Init. + Update
& \textbf{9.9} & \textbf{4.2} & \textbf{78.9}\\
\bottomrule
\end{tabular}
\caption{Comparison of memory construction protocols.}
\label{tab:memory_construction}
\end{table}

As shown in Table~\ref{tab:memory_construction}, the default
\textit{Online Persistent} memory outperforms the frozen pre-rendered
memory despite starting from an empty SOKB. Specifically, online
construction improves SR from 8.7 to 9.6 and SDTW from 3.4 to 4.0,
while reducing NE by 8.2\,m. This result indicates that the usefulness
of global memory depends not only on scene coverage, but also on which objects are retained and whether their representations
can be refined during navigation. The frozen initialization provides
broader scene-level coverage from the beginning, but its observations
are navigation-agnostic and therefore may include objects that are
visually distinctive yet irrelevant to the instructions encountered
during evaluation. Moreover, once constructed, erroneous or weakly
observed instances cannot be corrected.

In contrast, AirAnchor constructs object anchors conditioned on the
active subtask, navigation progress, and current plan. Consequently,
the objects entering the SOKB are biased toward landmarks and spatial
references that are directly useful for navigation. Repeated
observations from subsequent episodes further refine their geometry,
semantic descriptions, and visual evidence through the same
coarse-to-fine association mechanism. The simultaneous improvement in
SDTW and NE therefore suggests that online memory contributes not only
to identifying instruction-relevant landmarks, but also to providing
more useful spatial priors for maintaining accurate long-horizon
navigation.

The \textit{Pre-rendered Init. + Online Update} setting achieves the
best overall performance, but its gain over \textit{Online Persistent}
is comparatively small: SR increases by only 0.3 points, SDTW by 0.2,
and NE decreases by 1.6\,m. This pattern suggests that offline
initialization is mainly beneficial during the cold-start stage, when
few task-relevant landmarks have yet been accumulated. Once online
navigation experience provides sufficient coverage, continued
context-conditioned updates become the dominant source of useful
memory information. Therefore, the results support the central design
of AirAnchor: global spatial knowledge is most effective when it is
progressively constructed from the same task-relevant object anchors
used for local spatial grounding, rather than maintained as a static
scene representation.

\begin{table*}[ht]
\centering
\small
\setlength{\tabcolsep}{8.5pt}
\renewcommand{\arraystretch}{1.08}
\begin{tabular}{@{}lccccc@{}}
\toprule
\textbf{Method}
& \textbf{Calls}$\downarrow$
& \textbf{Input / Call}$\downarrow$
& \textbf{Output / Call}$\downarrow$
& \textbf{Total Tokens}$\downarrow$
& \textbf{SR}$\uparrow$ \\
\midrule
SPF
& \textbf{30.3}
& \textbf{681}
& \textbf{36}
& \textbf{21,725}
& 5.7 \\
FineCog-Nav
& 513.4
& 691
& 70
& 390,697
& 6.9 \\
AirAnchor
& 45.8
& 1,418
& 68
& 68,059
& \textbf{9.6} \\
\bottomrule
\end{tabular}
\caption{
Comparison of MLLM workload. \textit{Calls} denotes the average number of MLLM invocations per episode; \textit{Input / Call} and \textit{Output / Call} indicate the average numbers of input and output tokens per invocation, respectively; and \textit{Total Tokens} denotes the average token consumption per episode.
}
\label{tab:mllm_efficiency}
\end{table*}

\subsubsection{Robustness to Episode Ordering}
\label{app:episode_order}

The default AirAnchor evaluation follows the canonical episode order
provided by AerialVLN-S. To examine whether scene-persistent memory
depends on this ordering, we additionally evaluate three independent
random permutations of all Validation Seen episodes. For each run,
all scene memories are initialized as empty, and each episode can access
only the SOKB accumulated from earlier episodes of the same scene under
that permutation. All other settings are unchanged.

\begin{table}[t]
\centering
\small
\setlength{\tabcolsep}{8.0pt}
\renewcommand{\arraystretch}{1.08}
\begin{tabular}{@{}lccc@{}}
\toprule
\textbf{Episode Order}
& \textbf{SR}$\uparrow$
& \textbf{SDTW}$\uparrow$
& \textbf{NE}$\downarrow$\\
\midrule
Canonical
& 9.6
& 4.0
& 80.5\\
Random Shuffle
& 9.5 $\pm$ 0.2
& 4.0 $\pm$ 0.1
& 82.4 $\pm$ 3.8\\
\bottomrule
\end{tabular}
\caption{
Robustness to episode ordering. Random Shuffle reports mean and standard deviation over three independent episode permutations.
}
\label{tab:episode_order}
\end{table}

As shown in Table~\ref{tab:episode_order}, randomizing the episode
order causes little change in SR and SDTW, which remain close to the
canonical results. This indicates that the benefit of persistent memory
does not rely on a favorable evaluation sequence. The larger variation
in NE reflects the expected cold-start effect: changing the order
changes which same-scene landmarks have already been accumulated when
an episode is encountered, thereby affecting the accuracy of long-range
spatial priors and final approach.

This limited order sensitivity follows from AirAnchor's object-centric
memory construction. Object anchors are associated through semantic,
visual, and geometric consistency rather than episode identity, allowing
repeated landmarks observed from different trajectories to be
consolidated regardless of when they are encountered. As same-scene
experience accumulates, different episode orders therefore recover
largely overlapping task-relevant spatial knowledge. Overall, episode
ordering mainly affects when useful landmark priors become available,
rather than the persistent-memory benefit itself.

\subsubsection{MLLM Efficiency Analysis}
\label{app:mllm_efficiency}

Table~\ref{tab:mllm_efficiency} reveals different
efficiency--performance characteristics among the three methods.
SPF has the lowest MLLM workload, requiring only 30.3 calls and
21,725 tokens per episode. This follows directly from its lightweight
reactive design: each invocation maps the current visual observation
to a single spatial waypoint, with little additional structured
reasoning or persistent state involved. Such a compact control loop is
highly efficient, but each prediction is primarily grounded in the
current observation. Consequently, long-horizon instruction following
and spatial relations that extend beyond the current field of view
receive limited explicit support, which is reflected in its lower
SR of 5.7.

FineCog-Nav exhibits the opposite behavior. Although its average
per-call input size is only 691 tokens, close to that of SPF, its
primitive-action cognitive loop repeatedly invokes perception,
reasoning, decision, and memory-related modules throughout navigation.
As a result, the workload accumulates to 513.4 calls and 390,697 tokens
per episode. This dense cognitive processing provides frequent
closed-loop reassessment, but also repeatedly reprocesses closely
related visual and textual context at a fine control granularity.
The resulting computational overhead is therefore driven primarily by
the \emph{frequency} of MLLM reasoning rather than by unusually large
individual prompts.

AirAnchor occupies a different operating point. Its average input per
call is larger (1,418 tokens) because a high-level decision explicitly
incorporates compact local spatial anchors, persistent landmark priors,
and spatial evidence for progress reflection. However, these richer
reasoning calls are performed at the skill level and are amortized over
multiple primitive actions. AirAnchor therefore requires only 45.8
calls and 68,059 tokens per episode---substantially closer to SPF than
to FineCog-Nav in invocation frequency---while achieving the highest
SR of 9.6. Relative to FineCog-Nav, it reduces MLLM invocations by
$91.1\%$ and total token consumption by $82.6\%$.

These results highlight the efficiency advantage of AirAnchor's design:
rather than minimizing reasoning as in a purely reactive waypoint
policy, or repeatedly invoking multiple cognitive modules at every
primitive action, AirAnchor concentrates MLLM computation on a small
number of spatially informed high-level decisions. The anchor-based representation further allows local geometry and persistent global
knowledge to be communicated in a compact, structured form. This
enables substantially stronger navigation performance without the
dense MLLM interaction required by primitive-action cognitive agents.


\begin{figure*}[t]
  \includegraphics[width=\linewidth]{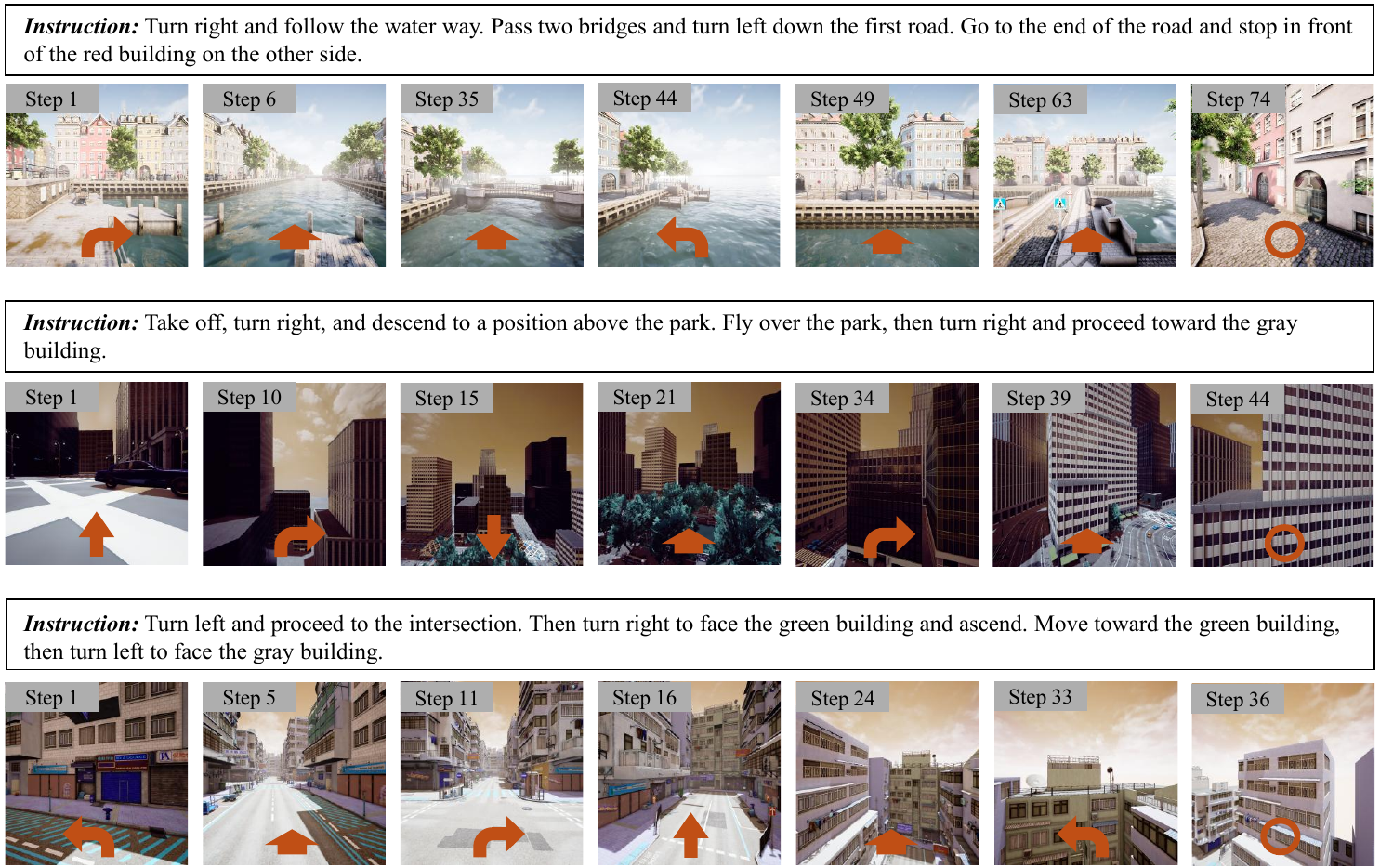}
  \caption{Visualization of key navigation steps from additional successful trajectories across different scenes.}
  \label{fig:add_case}
\end{figure*}

\subsection{Additional Qualitative Analysis}
\label{app:qualitative}

\subsubsection{Additional Successful Cases}
Additional successful trajectories from different scenes are shown in Figure~\ref{fig:add_case}. These cases demonstrate that AirAnchor can successfully navigate across diverse scene types and instruction conditions, highlighting the generalization and robustness of our method.

\subsubsection{Failure Modes}
\label{app:failure_modes}

Despite the benefits of cross-scale spatial reasoning, AirAnchor retains three limitations inherent to its design.

\noindent\textbf{Sparse grounding bottleneck.}
AirAnchor grounds only a small set of decision-relevant anchors.
If a critical landmark or motion direction is omitted during anchor
selection, its geometry is unavailable to all subsequent reasoning in
that decision step. This limitation is amplified at long range, where
objects with unreliable depth are degraded to directional anchors.
Although this prevents inaccurate geometry from entering the SOKB, it
also temporarily removes object identity and leaves the agent with only
coarse directional evidence.

\noindent\textbf{Persistent memory error propagation.}
The SOKB reuses accumulated object anchors across navigation episodes,
making association errors more consequential than transient perception
errors. Visually similar urban structures may be incorrectly merged,
while strong viewpoint changes may split one landmark into multiple
instances; either case can bias later landmark retrieval. Moreover,
landmarks that have not yet been observed provide no global prior.
Thus, the effectiveness of persistent spatial memory depends jointly
on reliable object association and sufficient online scene coverage.

\noindent\textbf{Ambiguous spatial completion.}
Progress reflection compares the same anchors before and after skill
execution, but linguistic relations such as ``near'', ``past'',
``across'', and ``around'' do not correspond to fixed geometric
thresholds. Incomplete or inaccurate anchor evidence can therefore
cause premature or delayed subtask completion. Since completion
directly controls subtask transition and the final \textsc{Stop}
decision, such errors can propagate into substantial trajectory
deviations near the destination.

\section{Prompt Templates}
\label{app:prompts}

The following templates specify the task instructions provided to the
MLLM in each AirAnchor module. Runtime observations and structured
navigation states are inserted into the corresponding placeholders.

\subsection{Instruction Decomposition Prompt}

\begin{promptbox}

\promptsec{Role.}

Decompose an aerial navigation instruction into an ordered sequence of
executable subtasks, extract the landmarks referenced by each subtask,
and further decompose each subtask into ordered subgoals solely for
computing its automatic-backtracking threshold.

\promptsec{Input.}

Navigation instruction:

\placeholder{INSTRUCTION}

\promptsec{Definitions.}

A \textbf{subtask} is a coherent segment of the instruction describing
an intermediate navigation objective. The ordered subtasks must
collectively preserve the complete execution semantics of the original
instruction.

A \textbf{subgoal} is a minimal instruction-supported navigation objective
within a subtask. Return at least one subgoal per subtask, preserving the
original action order and spatial relations without adding low-level actions
or arbitrarily splitting a single objective. Only the number of subgoals is
used by the controller, which sets the automatic-backtracking threshold to
three times that number. The complete subtask text remains the navigation
MLLM input; these auxiliary subgoals are not used to initialize the plan.

A \textbf{landmark} is a persistent scene object explicitly referenced
by the instruction. Its \texttt{category} describes the intrinsic
object type, while \texttt{attributes} contain only intrinsic visual
properties explicitly stated in the instruction, such as color,
material, shape, or structural appearance.

Spatial or temporal expressions such as left, right, above, below,
before, after, behind, across, around, and pass describe navigation
relations. They must remain in the subtask text and must not be encoded
as landmark attributes.

\promptsec{Reasoning Procedure.}

1. Read the complete instruction and identify its ordered navigation
   objectives.

2. Group actions that jointly describe one coherent intermediate
   objective into the same subtask while preserving their original
   execution order.

3. For each subtask, identify every explicitly referenced landmark and
   extract its category and stated intrinsic appearance attributes.

4. Preserve all spatial and temporal relations required to execute the
   instruction correctly.

5. Further decompose each subtask into a non-empty ordered list of
   minimal subgoals. Keep the complete subtask text unchanged and preserve
   all of its navigation requirements in the subgoal list.

6. Do not introduce objects, attributes, actions, or relations that are
   not supported by the instruction.

\promptsec{Output.}

Return JSON only:

\begin{promptcode}
{
  "subtasks": [
    {
      "id": "S1",
      "text": "...",
      "subgoals": ["..."],
      "landmarks": [
        {
          "id": "L1",
          "category": "...",
          "attributes": ["..."]
        }
      ]
    }
  ]
}
\end{promptcode}

If no intrinsic appearance attribute is explicitly given for a
landmark, return an empty \texttt{attributes} list.

\end{promptbox}

\subsection{Spatial Anchor Query Prompt}

\begin{promptbox}

\promptsec{Role.}

Select three visual references whose metric spatial information would
be most useful for the UAV's next navigation decision.

\promptsec{Inputs.}

Current $512\times512$ egocentric RGB observation:

\placeholder{IMAGE}

Current subtask:

\placeholder{SUBTASK}

Current navigation progress:

\placeholder{PROGRESS}

Current next-step plan:

\placeholder{PLAN}

\promptsec{Input Meaning.}

The \textbf{current subtask} specifies the active instruction objective.

The \textbf{navigation progress} summarizes which parts of that subtask
have already been achieved.

The \textbf{next-step plan} specifies the immediate objective that the
next navigation decision should accomplish.

The RGB observation is the UAV's current forward-facing view.

\promptsec{Anchor Types.}

\textbf{Object anchor.}
A clearly visible and spatially persistent object whose relative
location would help execute the current subtask.

For an object anchor, return only its intrinsic object
\texttt{category} and visible \texttt{attributes}. Object localization,
segmentation, confidence, distance, bearing, and height are obtained by
the external grounding and depth modules and must not be estimated
here.

\textbf{Directional anchor.}
An image direction whose depth-supported range and relative geometry
would help determine the next movement.

Represent a directional anchor by an integer pixel coordinate
\texttt{[u,v]}. The image origin is at the top-left;
$u\in[0,511]$ increases to the right and
$v\in[0,511]$ increases downward.

\promptsec{Reasoning Procedure.}

1. Identify the immediate action implied by the current subtask,
   progress, and next-step plan.

2. Determine which object locations or movement directions require
   explicit metric information to resolve that action.

3. Select exactly three anchors that provide the most useful and
   complementary spatial evidence.

4. Prefer instruction-relevant objects and decision-relevant directions
   over visually salient but navigation-irrelevant content.

5. Avoid redundant anchors that would provide nearly identical spatial
   information.

6. Do not infer metric distance, bearing, or height directly from RGB.

\promptsec{Output.}

Return JSON only:

\begin{promptcode}
{
  "anchors": [
    {
      "id": "A1",
      "type": "object",
      "category": "building",
      "attributes": ["gray", "rectangular"]
    },
    {
      "id": "A2",
      "type": "direction",
      "pixel": [u, v]
    }
  ]
}
\end{promptcode}

The \texttt{anchors} list must contain exactly three entries.
Any combination of object and directional anchors is allowed.

\end{promptbox}

\subsection{Landmark Prior Selection Prompt}

\begin{promptbox}

\promptsec{Role.}

Resolve the landmarks referenced by the current subtask against
candidate object instances retrieved from persistent scene memory.

\promptsec{Inputs.}

Current subtask:

\placeholder{SUBTASK}

Referenced landmarks:

\placeholder{LANDMARKS}

Retrieved candidate sets:

\placeholder{CANDIDATE_SETS}

\promptsec{Input Meaning.}

Each referenced landmark has a unique \texttt{landmark\_id}, an object
category, and any appearance attributes specified by the instruction.

The candidate set for a landmark contains persistent object instances
that have already passed coarse semantic and spatial retrieval.
Each candidate contains:

\begin{itemize}
    \item a unique candidate and persistent instance ID;
    \item stored category and appearance descriptions;
    \item a memory reliability score;
    \item its bearing relative to the current UAV;
    \item whether and how far it lies above or below the UAV;
    \item horizontal distance to the UAV;
    \item 3D distance to the UAV.
\end{itemize}

These spatial quantities are computed from the stored object geometry
and the current UAV pose. They should be used as metric spatial
evidence, while the memory reliability indicates how strongly the
stored instance is supported by previous observations.

\promptsec{Reasoning Procedure.}

For each referenced landmark:

1. Compare its category and appearance description with those of its
   candidate instances.

2. Interpret the spatial relations expressed by the current subtask and
   determine what approximate landmark configuration those relations
   imply.

3. Compare each candidate's UAV-relative spatial information with the
   configuration implied by the instruction.

4. When several landmarks occur in the same subtask, jointly use their
   relative ordering and spatial constraints to disambiguate otherwise
   similar candidates.

5. Use memory reliability only as supporting evidence; do not choose an
   instance solely because it has the highest reliability.

6. Select at most one candidate for each landmark. If the semantic and
   spatial evidence is insufficient or contradictory, return
   \texttt{null} rather than forcing a match.

\promptsec{Output.}

Return JSON only:

\begin{promptcode}
{
  "matches": [
    {
      "landmark_id": "L1",
      "instance_id": "O3",
      "reason": "concise semantic and spatial evidence"
    },
    {
      "landmark_id": "L2",
      "instance_id": null,
      "reason": "insufficient evidence"
    }
  ]
}
\end{promptcode}

Return one entry for every referenced landmark and preserve their
input order.

\end{promptbox}

\subsection{Skill-Level Navigation Prompt}

\begin{promptbox}

\promptsec{Role.}

Select one high-level navigation skill and its required parameters to
best advance the current aerial-navigation subtask.

\promptsec{Inputs.}

Current $512\times512$ RGB observation with anchor indices overlaid:

\placeholder{IMAGE}

Current subtask:

\placeholder{SUBTASK}

Current navigation progress:

\placeholder{PROGRESS}

Current next-step plan:

\placeholder{PLAN}

Local Egocentric Anchor Graph (EAG):

\placeholder{EAG}

Global landmark priors:

\placeholder{GLOBAL_PRIOR}

\promptsec{Navigation-State Meaning.}

The \textbf{current subtask} specifies the active instruction
objective. The \textbf{navigation progress} summarizes what has already
been achieved, and the \textbf{next-step plan} specifies the immediate
objective for the next action.

The \textbf{Egocentric Anchor Graph (EAG)} is a compact local spatial
representation constructed from the current RGB-D observation.
It contains the UAV and the spatial anchors selected for the current
decision. Each anchor is associated with the same numbered marker in
the RGB observation.

An \textbf{object anchor} represents a visible object and contains its
semantic description. Its spatial relation to the UAV includes relative
bearing, relative height, horizontal distance, and 3D distance.

A \textbf{directional anchor} represents a queried image direction.
Its spatial information describes the relative direction and the
depth-supported visible range along that camera ray. If the measured
range exceeds the reliable depth limit, the anchor explicitly indicates
that the reference has been capped rather than treating the capped
point as a physical surface.

The \textbf{global landmark priors} contain persistent memory instances
selected as likely matches for landmarks referenced by the current
subtask. Their positions are recomputed relative to the current UAV
pose. A prior can therefore provide directional and distance information
about an instruction-relevant landmark even when that landmark is
outside the current field of view.

Use the explicit metric spatial information in the EAG and global
priors rather than re-estimating precise geometry from RGB.

\promptsec{Available Skills.}

\textbf{Pixel Navigation}

Move toward an image-space direction visible in the current RGB
observation.

Parameters:

\begin{promptcode}
{"pixel":[u,v], "distance_m":d}
\end{promptcode}

The pixel must satisfy
$u,v\in[0,511]$. It may be any pixel in the current image and does not
need to coincide with an anchor marker.

The intended travel distance must satisfy
$5\leq d\leq50$ meters.

Use this skill when the required translational direction can be
determined from the current view.

\textbf{Altitude Adjustment}

Perform an explicit vertical displacement while maintaining the
current horizontal position.

Parameter:

\begin{promptcode}
{"delta_h_m":h}
\end{promptcode}

Positive values mean ascending and negative values mean descending,
with $|h|\leq30$ meters.

Use this skill when the instruction or current geometry requires a
change in altitude before further navigation.

\textbf{View Rotation}

Reorient the UAV when the required navigation direction cannot be
determined reliably from the current field of view.

Return an empty parameter object. Eight observations covering the
surrounding panorama will subsequently be acquired at $45^\circ$
intervals, and a dedicated panoramic reasoning module will infer a
fine-grained target yaw that is not restricted to the sampled viewing
angles.

Use this skill when an instruction-relevant landmark, expected route,
or required heading lies outside the current view or remains
directionally ambiguous.

\textbf{Path Backtracking}

Return along the previously executed path of the current subtask.

Return an empty parameter object. A dedicated history reasoning module
will subsequently select the previous path node to which the UAV
should return.

Use this skill only when the available evidence indicates that previous
navigation has deviated from the intended subtask and continuing from
the current state is unlikely to recover efficiently.

\promptsec{Reasoning Procedure.}

1. Identify the immediate navigation objective from the current
   subtask, progress, and next-step plan.

2. Use global landmark priors, when available, to determine the
   long-range landmark or direction relevant to that objective.

3. Use the current RGB observation and EAG to determine what movement
   is supported by the currently observed local geometry.

4. Decide whether the objective requires translation, vertical
   adjustment, additional viewing coverage, or recovery from a previous
   navigation error.

5. Select exactly one skill that most directly advances the immediate
   objective.

6. If Pixel Navigation or Altitude Adjustment is selected, determine
   the corresponding metric parameters from the available spatial
   evidence.

7. Do not determine whether the current subtask is complete. Completion
   is evaluated only after skill execution by the progress-reflection
   module.

Also produce a concise \texttt{scene\_caption} describing only the
navigation-relevant visual state. It is stored in the current-subtask
history for possible backtracking.

\promptsec{Output.}

Return JSON only:

\begin{promptcode}
{
  "skill": "Pixel Navigation",
  "parameters": {
    "pixel": [u, v],
    "distance_m": d
  },
  "reason": "concise spatial evidence for the selected skill",
  "scene_caption": "concise navigation-relevant scene description"
}
\end{promptcode}

The \texttt{skill} field must be exactly one of
\texttt{"Pixel Navigation"},
\texttt{"Altitude Adjustment"},
\texttt{"View Rotation"}, or
\texttt{"Path Backtracking"}.

For View Rotation and Path Backtracking, return
\texttt{"parameters": \{\}}.

\end{promptbox}

\subsection{Panoramic View Reorientation Prompt}

\begin{promptbox}

\promptsec{Role.}

Determine the concrete yaw rotation that best reorients the UAV for
continuing the current navigation subtask.

\promptsec{Inputs.}

Current subtask:

\placeholder{SUBTASK}

Current navigation progress:

\placeholder{PROGRESS}

Current next-step plan:

\placeholder{PLAN}

Global landmark priors:

\placeholder{GLOBAL_PRIOR}

Eight RGB observations captured at the same UAV position:

\placeholder{VIEW_0},
\placeholder{VIEW_1},
\ldots,
\placeholder{VIEW_7}

\promptsec{Input Meaning.}

The \textbf{current subtask} specifies the active navigation objective.
The \textbf{progress} summarizes what has already been achieved, and
the \textbf{next-step plan} describes the immediate objective that the
new heading should support.

The \textbf{global landmark priors} describe instruction-relevant
landmarks retrieved from persistent memory. When available, they
provide each landmark's current relative direction, relative height,
and distance from the UAV and can therefore help determine which
panoramic region should contain the desired route or landmark.

\promptsec{Panoramic View Definition.}

The eight images are captured at the same UAV position and jointly form
a sparse $360^\circ$ panoramic observation. Their yaw offsets are
defined relative to the heading before the panoramic scan:

\begin{promptcode}
VIEW_0:    0 deg
VIEW_1:   45 deg
VIEW_2:   90 deg
VIEW_3:  135 deg
VIEW_4:  180 deg
VIEW_5: -135 deg
VIEW_6:  -90 deg
VIEW_7:  -45 deg
\end{promptcode}

Positive yaw means turning right and negative yaw means turning left.
The eight views are visual references, not discrete action choices.
The desired heading may lie between two sampled views.

\promptsec{Turning-Direction Definition.}

Use the following coarse directional regions:

\begin{itemize}

    \item \textbf{Right}: a target heading on the right side,
    with
    $0^\circ < \texttt{yaw\_delta\_deg}\leq135^\circ$.

    \item \textbf{Left}: a target heading on the left side,
    with
    $-135^\circ\leq\texttt{yaw\_delta\_deg}<0^\circ$.

    \item \textbf{Around}: a target heading primarily behind the UAV,
    with
    $135^\circ<|\texttt{yaw\_delta\_deg}|\leq180^\circ$.
    Either rotation direction may be used according to which better
    aligns with the desired heading.

\end{itemize}

The final yaw is not restricted to multiples of $45^\circ$ or to the
eight sampled view directions.

\promptsec{Reasoning Procedure.}

1. Identify the heading required to advance the current subtask from
   the current progress and next-step plan.

2. Use any global landmark prior to estimate the coarse direction of
   the relevant landmark or route.

3. Examine all eight views jointly and identify the panoramic region
   that best supports the required heading.

4. Classify the required reorientation as
   \texttt{left}, \texttt{right}, or \texttt{around} using the angular
   definitions above.

5. Within that region, compare neighboring views and infer the
   fine-grained heading. The target may lie between the sampled yaw
   directions.

6. Output a signed yaw rotation in
   $[-180^\circ,180^\circ]$ relative to the heading before the scan.
   Choose the smallest rotation that correctly aligns the UAV with the
   immediate navigation objective.

\promptsec{Output.}

Return JSON only:

\begin{promptcode}
{
  "turning_direction": "right",
  "yaw_delta_deg": 70,
  "reason": "concise visual and spatial evidence for the target heading"
}
\end{promptcode}

The \texttt{turning\_direction} field must be one of
\texttt{"left"}, \texttt{"right"}, or \texttt{"around"} and must be
consistent with the sign and magnitude of
\texttt{yaw\_delta\_deg}.

\end{promptbox}

\subsection{Backtracking Node Selection Prompt}

\begin{promptbox}

\promptsec{Role.}

Select the previous path-history node to which the UAV should return
after agent-selected recovery or an automatic subtask-step trigger.

\promptsec{Inputs.}

Current subtask:

\placeholder{SUBTASK}

Current-subtask path history:

\placeholder{HISTORY_CHAIN}

\promptsec{History Definition.}

The path history contains only states and executed transitions from the
current subtask and is ordered chronologically.

Each \textbf{history node} contains:

\begin{itemize}
    \item a unique node ID;
    \item the UAV pose at that decision state;
    \item the navigation state recorded at that time;
    \item a concise scene caption;
    \item object-anchor spatial information observed at that state.
\end{itemize}

The object-anchor information records visible persistent objects and
their relative bearing, relative height, horizontal distance, and 3D
distance to the UAV at that history node.

Each \textbf{history transition} records the navigation skill executed
from the preceding node, the reason for that decision, and the actual
UAV displacement produced by the execution.

Directional anchors are not retained in the path history because they
are transient references associated with individual decisions.

\promptsec{Reasoning Procedure.}

1. Trace the history in chronological order and compare the recorded
   navigation states with the objective of the current subtask.

2. Identify the earliest transition after which the trajectory appears
   to have deviated from the intended objective.

3. Find the latest node immediately before that deviation where the
   subtask grounding and movement direction were still reliable.

4. Prefer the most recent valid recovery point rather than returning
   farther than necessary.

5. Select only a node ID that exists in the provided history.

\promptsec{Output.}

Return JSON only:

\begin{promptcode}
{
  "node_id": 4,
  "reason": "concise evidence identifying the last reliable state"
}
\end{promptcode}

\end{promptbox}

\subsection{Progress Reflection Prompt}

\begin{promptbox}

\promptsec{Role.}

Evaluate how the most recently executed navigation skill changed the
progress of the current subtask and determine whether that subtask is
complete.

\promptsec{Inputs.}

Current subtask:

\placeholder{SUBTASK}

Progress before execution:

\placeholder{PREVIOUS_PROGRESS}

Plan before execution:

\placeholder{PREVIOUS_PLAN}

RGB observation before execution:

\placeholder{IMAGE_BEFORE}

RGB observation after execution:

\placeholder{IMAGE_AFTER}

Executed skill and its parameters:

\placeholder{EXECUTED_SKILL}

Reason for selecting the skill:

\placeholder{DECISION_REASON}

Decision-time Egocentric Anchor Graph:

\placeholder{EAG_BEFORE}

Recentered Egocentric Anchor Graph:

\placeholder{RECENTERED_EAG}

\promptsec{Navigation-State Meaning.}

The \textbf{current subtask} specifies the active instruction
objective.

The \textbf{previous progress} summarizes which parts of that subtask
had already been achieved before the most recent skill was executed.

The \textbf{previous plan} describes the immediate navigation objective
that the executed skill was intended to accomplish.

The two RGB observations show the scene immediately before and after
execution of that skill.

\promptsec{EAG Definition.}

The \textbf{Egocentric Anchor Graph (EAG)} is a compact spatial
representation constructed at the decision state before skill
execution. It contains the UAV and the decision-relevant spatial
anchors grounded from the pre-execution RGB-D observation.

An \textbf{object anchor} represents a visible object. Its reference
point is the grounded object center.

A \textbf{directional anchor} represents a selected image direction.
Its reference point is obtained from the depth-supported point along
that camera ray. If its depth exceeds the reliable sensing range, the
reference is capped and explicitly marked as a far-range directional
reference rather than a physical surface.

For every anchor, its relation to the UAV contains:

\begin{itemize}
    \item relative bearing;
    \item relative height;
    \item horizontal distance;
    \item 3D distance.
\end{itemize}

The \textbf{decision-time EAG} records these relations before skill
execution.

The \textbf{recentered EAG} contains exactly the same anchor identities
and world reference points. It is \emph{not} a newly perceived or
re-detected graph. After the UAV moves, only each anchor's relation to
the new UAV pose is recomputed.

Therefore, changes between the decision-time and recentered EAGs
directly describe how the UAV moved relative to the same spatial
references.

\promptsec{Reasoning Procedure.}

1. From the current subtask, previous progress, and previous plan,
   identify what the executed skill was intended to achieve.

2. Check whether the executed skill and its parameters are consistent
   with that intended action.

3. Compare the before/after RGB observations to determine the semantic
   effect of the execution.

4. Compare each shared anchor between the decision-time and recentered
   EAGs. Use changes in bearing, relative height, horizontal distance,
   and 3D distance to determine the corresponding spatial effect.

5. For relational objectives such as approaching, passing, crossing,
   moving behind, or flying above/below a landmark, require the observed
   spatial changes to be consistent with that relation. Do not infer
   completion from visual similarity alone when explicit spatial
   evidence contradicts it.

6. Integrate the semantic and spatial evidence to update the current
   subtask progress.

7. Mark the subtask as \texttt{COMPLETED} only when all navigation
   conditions required by the current subtask are sufficiently
   supported. Otherwise mark it as \texttt{ONGOING}.

8. If the status is \texttt{ONGOING}, provide a concise next-step plan
   that addresses the remaining part of the current subtask.

9. If the status is \texttt{COMPLETED}, do not infer or plan the next
   subtask, because no next-subtask information is provided in this
   call. Return an empty \texttt{next\_plan}; the navigation controller
   will initialize the next subtask separately.

\promptsec{Output.}

Return JSON only:

\begin{promptcode}
{
  "progress": "concise updated description of current-subtask progress",
  "status": "ONGOING",
  "next_plan": "concise next action for the current subtask"
}
\end{promptcode}

The \texttt{status} field must be either
\texttt{"ONGOING"} or \texttt{"COMPLETED"}.

If \texttt{status} is \texttt{"COMPLETED"}, return:

\begin{promptcode}
"next_plan": ""
\end{promptcode}

\end{promptbox}

\subsection{Navigation-Agnostic Object Query Prompt}
\label{app:offline_prompt}

This prompt is used only for the pre-rendered memory-construction
experiment.

\begin{promptbox}

\promptsec{Role.}

Select objects from the current scene that are suitable for persistent
spatial landmark memory.

\promptsec{Input.}

Current $512\times512$ RGB observation:

\placeholder{IMAGE}

\promptsec{Landmark-Memory Criterion.}

A suitable memory object should be:

\begin{itemize}
    \item spatially persistent rather than transient;
    \item clearly visible in the current observation;
    \item sufficiently distinctive to support recognition from later
    aerial viewpoints.
\end{itemize}

Examples include distinctive buildings, bridges, towers, large
facilities, and other stable urban structures.

For each selected object, the \texttt{category} describes its intrinsic
object type and \texttt{attributes} describe only intrinsic visible
properties such as color, material, shape, or structural appearance.

Object localization, segmentation, confidence, and metric geometry are
computed by the same external grounding pipeline used for online
anchors and must not be estimated here.

\promptsec{Reasoning Procedure.}

1. Examine the current image for clearly visible persistent scene
   objects.

2. Prefer objects that are visually distinctive enough to be
   re-identified from different aerial viewpoints.

3. Avoid transient objects and generic structures when more distinctive
   persistent landmarks are available.

4. Select at most three objects.

5. Return only their semantic category and intrinsic visible
   attributes.

\promptsec{Output.}

Return JSON only:

\begin{promptcode}
{
  "objects": [
    {
      "category": "bridge",
      "attributes": ["gray", "arched"]
    }
  ]
}
\end{promptcode}

The \texttt{objects} list may contain zero to three entries.

\end{promptbox}

\end{document}